\documentclass{article}
\usepackage{iclr2027_conference,times}

\usepackage{amsmath,amsfonts,bm}

\def\eqref#1{equation~\ref{#1}}

\def\1{\bm{1}}

\DeclareMathAlphabet{\mathsfit}{\encodingdefault}{\sfdefault}{m}{sl}
\SetMathAlphabet{\mathsfit}{bold}{\encodingdefault}{\sfdefault}{bx}{n}

\newcommand{\rmmFrameSampPub}{44.51}
\newcommand{\rmmNoMemPub}{32.70}

\newcommand{\rmmFinalMean}{67.17}
\newcommand{\rmmFinalSd}{1.23}

\usepackage{hyperref}
\usepackage{url}
\usepackage{booktabs}
\usepackage{graphicx}
\usepackage{amsmath}
\usepackage{amssymb}
\usepackage{array}
\usepackage{tabularx}
\usepackage{multirow}
\usepackage{xcolor}

\definecolor{yzBlue}{RGB}{59,130,246}

\definecolor{yzOrange}{HTML}{C2410C}

\hypersetup{
  colorlinks=true,
  linkcolor=yzOrange,
  citecolor=yzBlue,
  urlcolor=yzBlue
}

\title{Simple Agentic Memory for Generalist Robot Policies}

\author{%
\begin{minipage}{\dimexpr\textwidth-2\tabcolsep\relax}
\centering
{\normalsize\bfseries
Yuyou Zhang\textsuperscript{1,*}\quad
Yunbei Zhang\textsuperscript{2,*}\quad
Miao Li\textsuperscript{1}\quad
Janet Wang\textsuperscript{2}}\\[4pt]
{\normalsize\bfseries
Zijian Jin\textsuperscript{3}\quad
Shilong Liu\textsuperscript{4,5}\quad
Ding Zhao\textsuperscript{1}}\\[6pt]
{\footnotesize\normalfont
\textsuperscript{1}Carnegie Mellon University\quad
\textsuperscript{2}Tulane University\quad
\textsuperscript{3}New York University\\[2pt]
\textsuperscript{4}Princeton University\quad
\textsuperscript{5}Columbia University}\\[15pt]
{\small\normalfont \textbf{Project Website: \url{https://simplearm.github.io/}}}
\end{minipage}}

\iclrfinalcopy
\let\SimpleARMBaseMaketitle\maketitle
\renewcommand{\maketitle}{%
  \SimpleARMBaseMaketitle
  \fancyhead{}%
  \lhead{Preprint}%
}

\begin{document}
\maketitle
\ificlrfinal
\begingroup
\renewcommand{\thefootnote}{*}
\footnotetext{Equal contribution.}
\endgroup
\fi

\begin{abstract}
Visual-memory systems commonly retain or compress past observations. Robot control additionally requires interaction-derived state that no individual frame may explicitly represent, such as persistent identity relations, accumulated progress, or ordered procedures. We introduce \emph{Simple Agentic Robot Memory} (\textbf{SimpleARM}), a training-free memory layer for frozen generalist robot policies. From the task instruction, SimpleARM specifies what to monitor; frozen perceptual tools maintain compact typed state online; structured access retrieves that state only when a proposed subgoal depends on history; and current-view grounding resolves recalled entities before execution. We evaluate SimpleARM on RoboMME, a benchmark of memory-dependent robot manipulation tasks that require history information no longer available in the current observation. Across all 16 tasks and three policy seeds, SimpleARM achieves \textbf{\rmmFinalMean\%} mean success, compared with \rmmFrameSampPub\% for the strongest non-oracle baseline. Matched ablations show mechanism specificity: removing relation, reference, progress, or route state produces large losses where the affected state is retrieved for control, while largely sparing other tasks. These results support a state-based view of robot memory: effective memory for control is not simply retained visual history, but compact task-relevant state derived from the interaction history.
\end{abstract}

\section{Introduction}
\label{sec:intro}

Vision--language--action (VLA) models are expanding robot policies from isolated skills toward
generalist control over diverse objects, embodiments, and language
instructions~\citep{brohan2023rt2,kim2025openvla,black2025pi0}. As instructions become more
compositional and tasks span longer horizons, success depends on more than interpreting the current view: an embodied
agent must remember what it observed, what its own actions changed, and what remains to be done.
Memory is therefore a core capability for embodied agents, just as context management and experience
accumulation are for language and multimodal
agents~\citep{packer2023memgpt,wang2023voyager,fan2025embodied}.

Long-video understanding offers increasingly capable ways to sample, compress, and retrieve visual history, while streaming-video models make this process causal and memory-bounded~\citep{wang2024videoagent,fan2024videoagent,zhang2024flashvstream,chen2024videollmonline,huang2025onlinevideo,liang2026oasis}. In both settings, history is commonly represented first as an observation sequence, and memory is organized around preserving useful visual evidence from that sequence. Yet RoboMME's own memory-budget study shows increasingly modest gains from allocating more tokens to perceptual history~\citep{dai2026robomme}, suggesting that capacity alone does not resolve the memory problem. Robot control instead requires asking not only which observations to retain, but which task-relevant state to maintain. Its object of memory is an interleaved history of observations and actions: actions change both the world and what the agent will observe next, so under partial observability the state sufficient for control must summarize past actions as well as observations~\citep{kaelbling1998planning}. Crucially, the information needed for the next decision may not be contained in any single frame. Robot memory is not exhausted by selecting a representative subset of frames. \emph{Robot memory is a problem of maintaining the task-relevant state induced by an action--observation history.}

This formulation has a cognitive analogue: human visual working memory is strongly selective, yet its retained content is not uniformly converted into symbols~\citep{vogel2005control,harrison2009decoding,kwak2022abstract}. This suggests a computational principle: preserve perceptual processing for the current scene while explicitly structuring the identity, progress, and trajectory variables that must persist across time.

RoboMME provides a concrete testbed for this principle through memory-dependent tasks for robotic generalist policies~\citep{dai2026robomme}. Many tasks exhibit \emph{perceptual aliasing}: the same current
observation can be compatible with different interaction histories and therefore require different
next actions~\citep{chrisman1992perceptual}. The robot may need to remember which container covered
an object, which block was manipulated in a demonstration, how many repetitions have been completed,
or which route was shown earlier. In each case, the task-relevant state is determined by the
interaction history but is not fully recoverable from selected frames alone.

We address this ambiguity with \textbf{SimpleARM}, a training-free memory layer around a frozen grounded-subgoal predictor and VLA policy. Before execution, an
instruction-conditioned VLM proposes a candidate state specification and compatible retrieval operations.
During interaction, frozen perceptual and event tools verify evidence and maintain
entity--relation bindings, event progress, and ordered trajectories or procedures.
Structured access retrieves only state compatible with a history-dependent field of the proposed
subgoal, and current-view perception re-grounds recalled entities before execution.
This division of labor defines the agentic design: language specifies \emph{what} may matter, tools estimate and
update \emph{what is true}, and structured access determines \emph{when} maintained state can affect control.

Our contributions are threefold:
\begin{itemize}

\item We formulate robot memory as task-conditioned maintenance of \emph{policy-sufficient state}
over an action--observation history, rather than visual-history compression alone.

\item We introduce \emph{Simple Agentic Robot Memory} (\textbf{SimpleARM}), which derives a
task-specific memory specification, maintains compact state online, and combines structured retrieval with current-view grounding to expose that state to a frozen policy.

\item Across all 16 RoboMME tasks, our method achieves
\textbf{\rmmFinalMean\%} mean success over three seeds, exceeding both the no-memory baseline (\rmmNoMemPub\%) and the strongest previously reported non-oracle method
(\rmmFrameSampPub\%).
Ablations show that removing relation,
reference, progress, or route state causes large losses where that state is retrieved for control,
while largely sparing other tasks.

\end{itemize}

\section{Related work}
\label{sec:related}

\subsection{Agentic and online video understanding}

Long-video agents turn recordings into query-relevant evidence through iterative frame search,
visual tools, instruction-conditioned memory, or hierarchical reasoning
~\citep{wang2024videoagent,fan2024videoagent,diko2025rewind,yin2026videoarm,yeo2026worldmm,zhou2025reagentv}.
The central problem is selecting evidence from a fixed visual history. Online video models instead
process frames causally and answer temporally aligned or asynchronous queries under bounded memory
~\citep{zhang2024flashvstream,chen2024videollmonline,
huang2025onlinevideo,liang2026oasis}. SimpleStream shows that a sliding window of recent frames can
be a strong streaming baseline, motivating an explicit distinction between recent-scene perception
and long-range memory~\citep{shen2026simplestream}. Recent systems further organize streams into sparse,
entity-centric, or embodied scene memories~\citep{jiang2026memento,long2026m3,fan2025embodied}.
Dynamic Hub-and-Spoke Memory (D-HSM) maintains entity-centered textual history and combines
question-adaptive retrieval with recent visual frames for training-free, frozen-VLM answer
prediction~\citep{jiang2026dynamic}.
Robot control shares this causal, bounded-memory setting but couples perception to action: each
action changes the scene and subsequent observations. Memory must therefore supply a
control-relevant variable and, for recalled entities, re-ground identity in the current view before
action rather than return historical evidence only as an answer.

\subsection{Memory for VLAs}

RoboMME~\citep{dai2026robomme}, MIKASA-Robo~\citep{cherepanov2025mikasa},
RMBench~\citep{chen2026rmbench}, and RoboMemArena~\citep{lei2026robomemarena}
evaluate complementary forms of history dependence and partial observability in robot
manipulation. Memory-augmented VLA methods likewise span several representations: MemoryVLA
consolidates perceptual and cognitive tokens~\citep{shi2025memoryvla}; VPWEM combines working and
episodic memory~\citep{lei2026vpwem}; MEM combines short-horizon visual and long-horizon textual
memory~\citep{torne2026mem}; Chameleon learns control-indexed episodic state
~\citep{guo2026chameleon}; and MemER retrieves task-relevant keyframes for high-level action
prediction~\citep{sridhar2026memer}.

Several systems connect semantic memory to visuomotor control through a hierarchical interface.
RoboMME calls its high-level VLM a \emph{subgoal predictor}, and its symbolic-memory variants encode
history as language subgoals~\citep{dai2026robomme}. MemER similarly retrieves keyframes to produce
text instructions for a low-level VLA~\citep{sridhar2026memer}. More broadly, $\pi_{0.5}$ uses
semantic subtask prediction for hierarchical inference~\citep{black2025pi05}, while Hi Robot uses a
high-level VLM to infer the next step for a low-level VLA~\citep{shi2025hirobot}. We follow this
terminology and call the semantic planner the \emph{high-level subgoal predictor}.

These methods differ in what history reaches the policy: compressed visual tokens, retrieved
frames, or text. RoboMME further reports diminishing returns from scaling perceptual-memory tokens
~\citep{dai2026robomme}. \textbf{SimpleARM} instead maintains instruction-conditioned
entity--relation, progress, and trajectory state, retrieves it only for compatible
history-dependent subgoals, and re-grounds recalled entities in the current view. This
training-free layer leaves both the high-level subgoal predictor and low-level VLA frozen.

\subsection{Selective memory for grounded action}

General agent memories study what to store, consolidate, and retrieve for later reasoning
~\citep{packer2023memgpt,xu2025amem,niu2026survey}. DPCore preserves domain-specific knowledge through a
dynamically updated prompt coreset for continual test-time adaptation~\citep{zhang2025dpcore}.
SimpleARM maintains episode-specific task state while keeping model parameters and prompt
templates fixed. Closed-loop control additionally requires state to track
action-induced scene changes and retrieval to affect the policy only when the current decision
depends on history. Contextual cues and reliability-aware evidence fusion also support visual
inference~\citep{wang2022automatically,xu2026rivatfuse}.
Human visual working memory offers a complementary analogy: it selectively
admits task-relevant objects~\citep{vogel2005control}, retains stimulus-specific content
~\citep{harrison2009decoding}, recodes content into goal- and action-oriented abstractions
~\citep{kwak2022abstract,boettcher2021output}, and organizes continuous experience into event
models~\citep{kurby2008segmentation}. Together, these findings motivate preserving current visual
processing while explicitly structuring only the identity, progress, and trajectory variables
needed for future action.

\section{Method}
\label{sec:method}

\subsection{Policy-Sufficient State}
\label{sec:policy_state}
\label{sec:overview}

Let $g$ denote the task instruction, $d$ an optional demonstration, $o_t$ the observation at time
$t$, and $a_t$ the executed action. The information available to the agent is the interaction
history
\begin{equation}
    h_t=(g,d,o_0,a_0,\ldots,a_{t-1},o_t).
    \label{eq:interaction_history}
\end{equation}
SimpleARM maintains an episode-scoped state through a task-conditioned recursive update,
\begin{equation}
    M_t=\mathcal U(M_{t-1},o_t,a_{t-1};g,d),
    \label{eq:memory_update_abstract}
\end{equation}
with the objective that the current observation and memory retain the information from $h_t$ needed
for control:
\begin{equation}
    \pi(a_t\mid h_t,g)\approx\pi(a_t\mid o_t,g,M_t).
    \label{eq:policy_sufficiency}
\end{equation}
The policy-sufficient memory state need not correspond to any subset of past observations. Rather
than preserving only visual evidence, it can represent derived interaction state: persistent
entity--relation bindings, accumulated event progress, and ordered trajectories or procedures.

\begin{figure}[t]
\centering
\includegraphics[width=\textwidth]{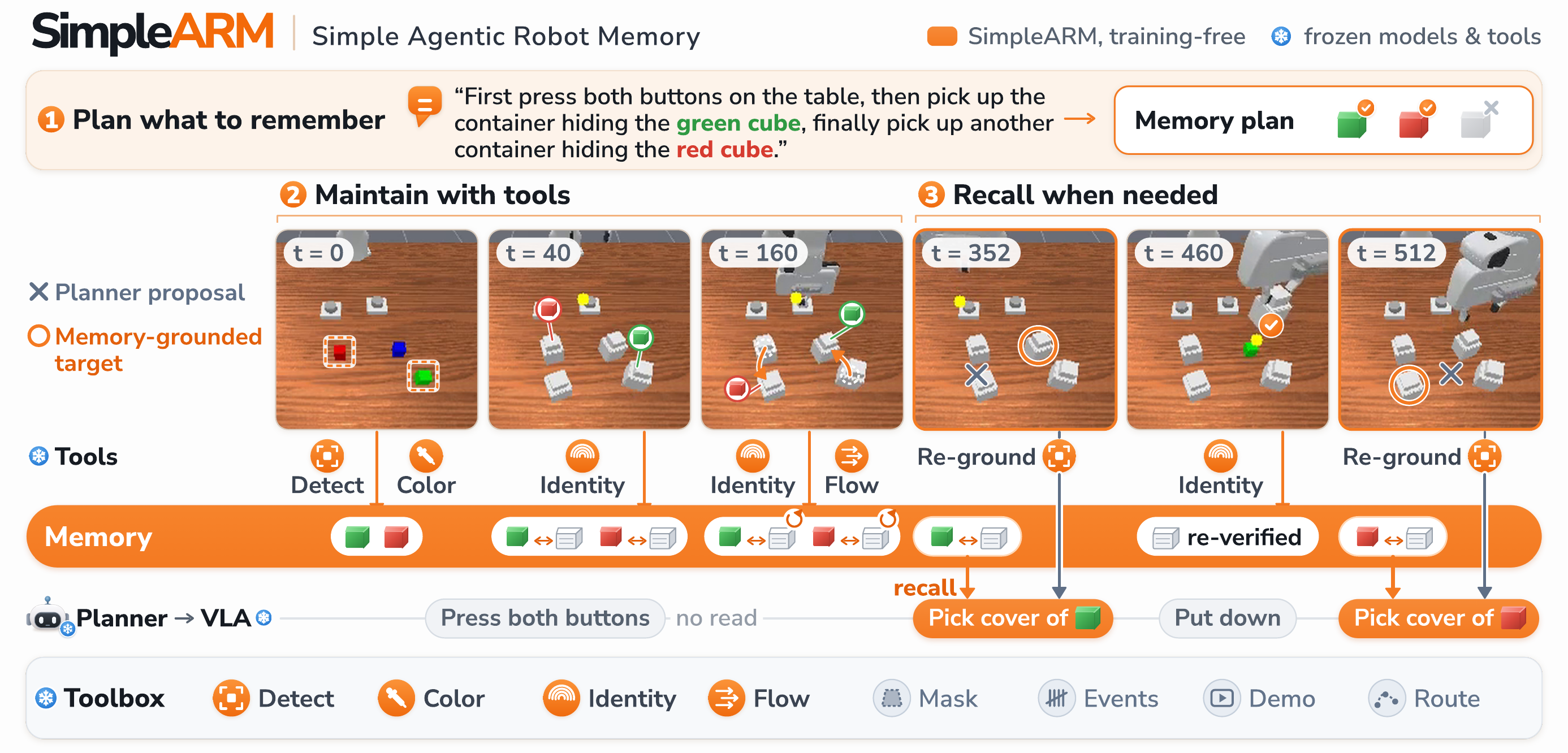}
\caption{\textbf{SimpleARM overview.}
Given the instruction, SimpleARM first specifies what historical state may be needed,
then maintains that state online with frozen perception and event tools, and retrieves it only
when the current subgoal depends on interaction history. Retrieved memory resolves historical
identity, progress, or procedure; current perception provides any required spatial grounding,
and the resulting memory-conditioned grounded subgoal is executed by the frozen VLA.}
\label{fig:overview}
\end{figure}

Figure~\ref{fig:overview} summarizes \textbf{SimpleARM}. Under hierarchical control, the
high-level subgoal predictor proposes the next subgoal and a frozen VLA policy executes it.
\textbf{SimpleARM} augments this pipeline in three stages: it specifies what historical
information may matter for future control, thereby restricting memory to task-relevant state;
maintains that state online using frozen perception and event tools, thereby preserving
reliable information across time; and selectively retrieves it only when the current subgoal
depends on interaction history, thereby limiting irrelevant memory influence on control.

Before execution, an instruction-conditioned memory specification identifies candidate state
variables to track. During interaction, frozen perceptual and event tools extract evidence from the
observation stream and update these variables. At decision time, structured access determines
whether the predicted subgoal depends on history; retrieved state is verified and, when necessary,
re-grounded in the current observation before being passed to the policy. Thus, language specifies
\emph{what} may matter, tools estimate and update \emph{what is true}, and structured access determines
\emph{when} maintained state can affect control.

Formally,
\begin{align}
    (P,A) &= \mathcal C(g),
    &M_t &= \mathcal U\!\left(
        M_{t-1},
        \Phi(o_t,a_{t-1},d);
        P
    \right),
    &\hat u_k &= \mathcal G_\theta(g,o_t),
    \label{eq:framework_write}\\
    r_k &= \mathcal R(\hat u_k,M_t;P,A),
    &\tilde u_k &= \mathcal V(\hat u_k,r_k,o_t),
    &a_{k:k+H} &\sim \pi_{\mathrm{VLA}}(\cdot\mid o_t,\tilde u_k).
    \label{eq:framework_read}
\end{align}
Here, $\mathcal C$ produces a candidate memory specification $P$ and compatible reader set $A$.
The tools $\Phi$ extract evidence from the current observation, executed action, and optional
demonstration.
$\mathcal U$ maintains the resulting typed state. We instantiate the abstract update in Eq.~\ref{eq:memory_update_abstract}
by first compiling a task-conditioned specification $P$ and extracting
observable evidence $\Phi(o_t,a_{t-1},d)$.
$\mathcal G_\theta$ is the frozen high-level subgoal predictor.
Given its proposed subgoal $\hat u_k$, $\mathcal R$ reads only compatible historical state, while
$\mathcal V$ verifies the retrieved information and, when needed, re-grounds referenced entities
in the current observation. The resulting subgoal $\tilde u_k$ is then executed by the frozen
VLA policy $\pi_{\mathrm{VLA}}$. SimpleARM separates three roles: the task instruction defines a
compact state abstraction, perceptual and event tools estimate that state from interaction, and structured access determines when it can influence control.

\subsection{Task-Conditioned State Abstraction}
\label{sec:state_maintenance}
\label{sec:write}

A complete interaction history contains redundant and distracting information for a single task. SimpleARM therefore constructs a task-conditioned abstraction of
that history: the instruction specifies a small set of state variables that may be
needed for future decisions, and perceptual and event tools estimate those variables
as interaction unfolds. Unlike a fixed latent representation, the state space itself
is configured by the task.

\paragraph{Task-conditioned state specification.}
Before execution, a task-conditioned memory specification produces a candidate state
specification
\begin{equation}
    P_g=\mathcal C(g)
    =\bigl(P_g^{\mathrm{ent}},P_g^{\mathrm{prog}},P_g^{\mathrm{traj}}\bigr),
    \label{eq:memory_specification}
\end{equation}
describing which persistent relations, progress variables, and ordered procedures may
be relevant to the task. The specification defines \emph{what should be estimated},
rather than its value: language proposes the state variables, while their values are
established from observations, actions, and an optional demonstration by frozen
perceptual and event tools.

The resulting episode state is
\begin{equation}
    M_t=
    \left(
        M_t^{\mathrm{ent}},
        M_t^{\mathrm{prog}},
        M_t^{\mathrm{traj}}
    \right),
    \label{eq:typed_memory}
\end{equation}
where the factors capture complementary forms of history-dependent information.
$M_t^{\mathrm{ent}}$ is an object-centric relational state that preserves identities and
task-relevant relations after their original visual evidence becomes unavailable;
$M_t^{\mathrm{prog}}$ represents task progress accumulated through interaction, such as event counts or repetition state; and $M_t^{\mathrm{traj}}$ represents ordered procedural state, such as a demonstrated route or sequence together with its current progress.
These variables summarize information induced by the interaction history rather than
retaining the observations themselves.
Their representation and update rules differ, but all are instantiated only when selected by the task-conditioned memory specification.
This factorization is not intended as a complete world model: it retains only variables selected as potentially relevant to future control.

\paragraph{Tool-grounded state updates.}
Each maintained variable is updated from evidence produced by an appropriate perceptual
or event tool. Let
\begin{equation}
    z_t^i=\Phi_i(o_t,a_{t-1},d)
\end{equation}
denote candidate evidence for state variable $m^i$. An update is committed only when the
corresponding evidence gate accepts that evidence:
\begin{equation}
m_t^i=
\begin{cases}
    \mathcal U_i(m_{t-1}^i,z_t^i), &
    \mathcal A_i(z_t^i,o_t)=1,\\
    m_{t-1}^i, & \text{otherwise}.
\end{cases}
\label{eq:verified_write}
\end{equation}
Here, $\Phi_i$ extracts evidence relevant to the maintained variable, while
$\mathcal A_i$ determines whether that evidence is sufficient to update persistent state.
Thus, the task-conditioned specification determines \emph{what} should be maintained,
while tool-grounded evidence determines \emph{when and how} that state changes.
Details are described in
Appendices~\ref{app:schemas} and~\ref{app:instantiation}.

\subsection{Selective Memory Retrieval and Current-View Grounding}
\label{sec:memory_access}
\label{sec:read}

The task-conditioned state in Sec.~\ref{sec:state_maintenance} may contain information that is
relevant somewhere in the episode but unnecessary for the current decision. SimpleARM
therefore retrieves memory selectively rather than exposing the full maintained state to the
policy. At decision step $k$, the frozen high-level subgoal predictor proposes
\begin{equation}
\hat u_k=\mathcal G_\theta(g,o_t).
\end{equation}
The instruction-conditioned retrieval specification $A$ then determines whether the proposed
subgoal contains a history-dependent argument that can be resolved from $M_t$. Let
$\mathcal Q$ construct the corresponding memory query:
\begin{equation}
q_k=\mathcal Q(\hat u_k;P,A),
\qquad
r_k=
\begin{cases}
\mathcal R(M_t,q_k), & q_k\neq\varnothing,\\
\varnothing, & q_k=\varnothing.
\end{cases}
\label{eq:selective_access}
\end{equation}
Thus, memory retrieval is conditioned jointly on the task specification and the current
subgoal. An entity or relation query retrieves the historical referent needed to resolve an
object argument; a progress query retrieves an accumulated count, ordinal, or task stage; and
a procedural query retrieves the relevant element of a maintained sequence. A retrieval
operation proposed from the instruction may therefore never fire if no compatible
history-dependent argument appears during execution.

Historical state may determine \emph{which} entity is relevant without specifying \emph{where}
it is in the current observation. We therefore separate memory retrieval from current-view
grounding:
\begin{equation}
\gamma_k=
\begin{cases}
\operatorname{Ground}(r_k,o_t),
& \text{if current-view localization is required},\\
\varnothing,
& \text{otherwise},
\end{cases}
\qquad
\tilde u_k=\mathcal V(\hat u_k,r_k,\gamma_k).
\label{eq:current_grounding}
\end{equation}
Entity and relation state may require current-view grounding, whereas progress and procedural
state can directly refine symbolic subgoal arguments.

Memory modifies the proposal only when compatible historical state is available and any required
current grounding succeeds; otherwise, $\tilde u_k=\hat u_k$. Together with Sec.~\ref{sec:state_maintenance},
this yields a simple pipeline: the instruction specifies what to maintain, online evidence updates
that state, and selective retrieval determines when it influences the current subgoal.
Appendix~\ref{app:qualitative_lifecycle} illustrates this process over a complete episode.

\subsection{Training-Free Integration}
\label{sec:integration}

SimpleARM introduces a training-free memory layer for maintaining and selectively retrieving task-relevant state, while keeping the pretrained subgoal predictor and VLA policy frozen.
The high-level subgoal predictor is RoboMME's fine-tuned
Qwen3-VL-4B-Instruct model~\citep{qwen3vl}, which predicts grounded subgoals from visual observations,
while the low-level controller is its $\pi_{0.5}$-based GroundSG VLA
policy~\citep{dai2026robomme,black2025pi05}. Both remain frozen.
The memory layer uses frozen Qwen3-VL-4B-Instruct as an instruction-only planner to derive two
specifications before control begins. The state specification $P$ identifies candidate entity
relations, progress signals, and procedures to maintain; the retrieval specification $A$ describes
how those records may resolve a future history-dependent subgoal. This planning step proposes the
memory schema but neither assigns state values nor decides at runtime when memory is read. Frozen
perception and event tools maintain the state from observations, structural access retrieves
compatible records, and current-view perception grounds recalled entities before execution. The
query structure and vote aggregation are described in Appendix~\ref{app:compiler}.
The added memory is non-parametric and episode-scoped: $M_0$ is empty, state evolves
only through the updates in Sec.~\ref{sec:state_maintenance}, and all records are reset
between episodes. Appendix~\ref{app:method_details} gives the complete RoboMME instantiation.

\section{Experiments}
\label{sec:experiments}

Our experiments ask three questions that follow the state-to-action decomposition of
Section~\ref{sec:method}. \textbf{RQ1} asks \emph{when} maintaining task state improves over
retaining visual history, comparing SimpleARM with published and episode-matched retained-frame
baselines at the aggregate, suite, and episode level. \textbf{RQ2} asks \emph{which} components
matter: matched ablations test whether different memory demands rely on different maintained
variables, and whether a variable helps only when structured access retrieves it for a decision.
\textbf{RQ3} asks \emph{what remains} beyond memory, examining representative failures in which the
outcome hinges on state correctness, identity ambiguity, temporal perception, or downstream control.

\subsection{Experimental Setup}
\label{sec:setup}

We evaluate all sixteen RoboMME tasks~\citep{dai2026robomme}, grouped into the official
\emph{Counting}, \emph{Permanence}, \emph{Reference}, and \emph{Imitation} suites, across three seeds.
We initialize the high-level subgoal predictor from RoboMME's fine-tuned Qwen3-VL-4B-Instruct
grounded-subgoal adapter and the VLA from its $\pi_{0.5}$-based GroundSG
policy.
Both are used without further
training; the perception models also remain frozen, and the added memory receives no training
or test-time optimization.
Baselines are RoboMME's published three-seed $\pi_{0.5}$ (no-memory) and
FrameSamp-Modul results and its non-deployable oracle-subgoal
ceiling~\citep{dai2026robomme}, together with RoboMME's
evaluation of MemER~\citep{sridhar2026memer,dai2026robomme}.
Motivated by SimpleStream's recent-frame baseline for streaming video~\citep{shen2026simplestream},
we also report Recent32+ModuL, RoboMME's FrameSamp+ModuL architecture trained and evaluated with the
latest 32 frames as memory, using the same three policy seeds as our method (50 episodes per task per
seed); the
Recent4 and Recent16 variants and full protocol details are given in Appendix~\ref{app:statistics}.

\subsection{RQ1: When Does State Maintenance Improve over Visual-History Retention?}
\label{sec:rq1}

SimpleARM achieves \textbf{\rmmFinalMean\%} average success
(Table~\ref{tab:main}), outperforming the strongest baseline FrameSamp result by
22.66 points and $\pi_{0.5}$ without memory by 34.5 points. Recent32+ModuL reaches $31.50\pm0.43$\% over the same
three policy seeds, 13.0 points below FrameSamp.
The episode-matched comparison shows a similar margin: on the same 800 episodes,
our method succeeds on 542, compared with 369 for the released FrameSamp-Modul
checkpoint (+21.6 points), with 254 ours-only versus 81 FrameSamp-only successes.
Together, these results indicate a substantial system-level advantage over
retained-frame memory, although this comparison alone does not isolate which gains are specifically attributable to the memory layer.

\begin{table}[t]
\caption{\textbf{Success rate (\%) on RoboMME.}
All columns use three policy seeds and 50 episodes per task per seed.
Published baseline and oracle results are from
RoboMME~\citep{dai2026robomme}, which also evaluates
MemER~\citep{sridhar2026memer}; Recent32+ModuL retains the latest 32 native frames
(Sec.~\ref{sec:setup}).
$\Delta_{\mathrm{FS}}$ is ours minus FrameSamp.
State labels are not mutually exclusive.
Bold and underlining mark the best and second-best non-oracle results, respectively; ties share the
same formatting. The ground-truth-subgoal
oracle is shown in gray.}
\label{tab:main}
\centering
\scriptsize
\setlength{\tabcolsep}{1.9pt}
\renewcommand{\arraystretch}{0.92}
\begin{tabular}{@{}cllrrrrrr@{}}
\toprule
& & & \multicolumn{6}{c}{success (\%)} \\
\cmidrule(l){4-9}
suite & task & required state & no mem. & MemER & FrameSamp & Recent32 & ours & {\color{gray}oracle} \\
\midrule
\multirow{5}{*}{\rotatebox[origin=c]{90}{\itshape Counting}}
& BinFill & progress & 52.0 & \underline{56.7} & 39.6 & 34.0 & \textbf{60.0} & {\color{gray}85.8} \\
& PickXtimes & progress & \textbf{92.7} & 79.3 & 87.3 & 83.3 & \underline{91.3} & {\color{gray}100.0} \\
& SwingXtimes & progress & 7.3 & 59.3 & \textbf{92.0} & \underline{80.7} & 76.7 & {\color{gray}100.0} \\
& StopCube & progress & 0.0 & 0.0 & \textbf{42.0} & \underline{41.3} & 0.0 & {\color{gray}49.7} \\
\cmidrule(lr){2-9}
& \textit{suite avg.} & $\Delta_{\mathrm{FS}}=-8.22$ & 38.00 & 48.83 & \textbf{65.22} & \underline{59.83} & 57.00 & {\color{gray}83.86} \\
\cmidrule(lr){2-9}
\addlinespace[1pt]
\multirow{5}{*}{\rotatebox[origin=c]{90}{\itshape Permanence}}
& VideoUnmask & relation & \underline{88.7} & 81.3 & 32.7 & 19.3 & \textbf{96.0} & {\color{gray}98.8} \\
& ButtonUnmask & relation & 24.0 & \underline{72.0} & 25.1 & 20.0 & \textbf{82.0} & {\color{gray}95.0} \\
& VideoUnmaskSwap & relation & 30.7 & \underline{38.0} & 24.4 & 23.3 & \textbf{80.0} & {\color{gray}99.2} \\
& ButtonUnmaskSwap & relation+progress & 14.0 & 21.3 & 18.2 & \underline{28.0} & \textbf{56.7} & {\color{gray}80.2} \\
\cmidrule(lr){2-9}
& \textit{suite avg.} & $\Delta_{\mathrm{FS}}=+53.56$ & 39.34 & \underline{53.17} & 25.11 & 22.67 & \textbf{78.67} & {\color{gray}93.31} \\
\cmidrule(lr){2-9}
\addlinespace[1pt]
\multirow{5}{*}{\rotatebox[origin=c]{90}{\itshape Reference}}
& PickHighlight & relation+progress & 15.3 & \textbf{70.7} & 22.9 & 26.7 & \underline{66.0} & {\color{gray}83.3} \\
& VideoRepick & identity+progress & 25.3 & 25.3 & \underline{30.4} & 26.7 & \textbf{78.0} & {\color{gray}97.3} \\
& VideoPlaceButton & identity+order & 54.0 & 30.0 & \underline{60.0} & 26.0 & \textbf{82.0} & {\color{gray}100.0} \\
& VideoPlaceOrder & identity+order & 31.8 & 26.0 & \underline{32.0} & 24.7 & \textbf{88.0} & {\color{gray}100.0} \\
\cmidrule(lr){2-9}
& \textit{suite avg.} & $\Delta_{\mathrm{FS}}=+42.17$ & 31.61 & \underline{38.00} & 36.33 & 26.00 & \textbf{78.50} & {\color{gray}95.17} \\
\cmidrule(lr){2-9}
\addlinespace[1pt]
\multirow{5}{*}{\rotatebox[origin=c]{90}{\itshape Imitation}}
& MoveCube & procedure & 71.6 & \textbf{82.7} & \underline{77.8} & 52.0 & 67.3 & {\color{gray}87.8} \\
& InsertPeg & procedure & 3.3 & \underline{6.7} & \textbf{7.6} & 2.0 & 4.0 & {\color{gray}15.6} \\
& PatternLock & trajectory+order & 6.7 & 16.7 & \underline{53.6} & 8.7 & \textbf{93.3} & {\color{gray}97.0} \\
& RouteStick & trajectory+order & 6.0 & 12.0 & \textbf{66.7} & 7.3 & \underline{53.3} & {\color{gray}55.6} \\
\cmidrule(lr){2-9}
& \textit{suite avg.} & $\Delta_{\mathrm{FS}}=+3.11$ & 21.89 & 29.50 & \underline{51.39} & 17.50 & \textbf{54.50} & {\color{gray}63.98} \\

\midrule
\multicolumn{2}{@{}l}{\textbf{Overall avg.}} & $\Delta_{\mathrm{FS}}=+22.66$ & 32.70 & 42.37 & \underline{44.51} & 31.50 & \textbf{67.17} &
{\color{gray}84.08} \\
\bottomrule
\end{tabular}
\end{table}

\begin{figure}[t]
\centering
\includegraphics[width=\textwidth]{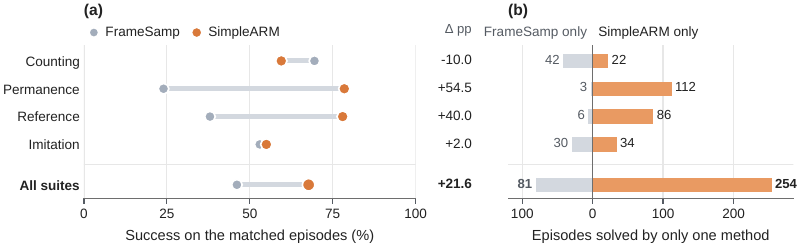}
\caption{\textbf{SimpleARM's paired gains are concentrated in Permanence and Reference.}
(a) Success rates of SimpleARM and the released FrameSamp-Modul checkpoint on matched episode
identities under the same seed (200 episodes per suite; 800 overall). $\Delta$ denotes the paired
success-rate difference (SimpleARM minus FrameSamp), in percentage points.
(b) Counts of episodes solved by exactly one system: FrameSamp alone (left) or SimpleARM alone
(right).
The bottom row in each panel pools all four suites.}
\label{fig:paired_suites}
\end{figure}

SimpleARM delivers its largest gains on \emph{Permanence} and \emph{Reference},
outperforming FrameSamp by 53.56 and 42.17 points, respectively
(Table~\ref{tab:main}). It also improves Imitation by 3.11 points. The only suite-level deficit is
Counting ($-8.22$ points), driven largely by the unsupported StopCube task; excluding it,
our method averages 76.0\% versus 73.0\% for FrameSamp on the remaining Counting tasks.

The episode-level comparison in Figure~\ref{fig:paired_suites} shows the same pattern: of the 115
Permanence episodes solved by exactly one system, ours solves 112, and of the 92 such Reference
episodes, ours solves 86.
Overall, the strongest gains arise when agentic tool use can accurately capture and retain
task-relevant historical state that is no longer available from the current observation or a
single retained frame. These system-level results do not isolate the effect of individual memory components. We therefore
ablate state formation, updates, and retrieval in Figure~\ref{fig:localization} and
Table~\ref{tab:final_ablation_matrix}.



\subsection{RQ2: Which Components of State Maintenance Matter?}
\label{sec:rq2}

\begin{figure}[t]
\centering
\includegraphics[width=\textwidth]{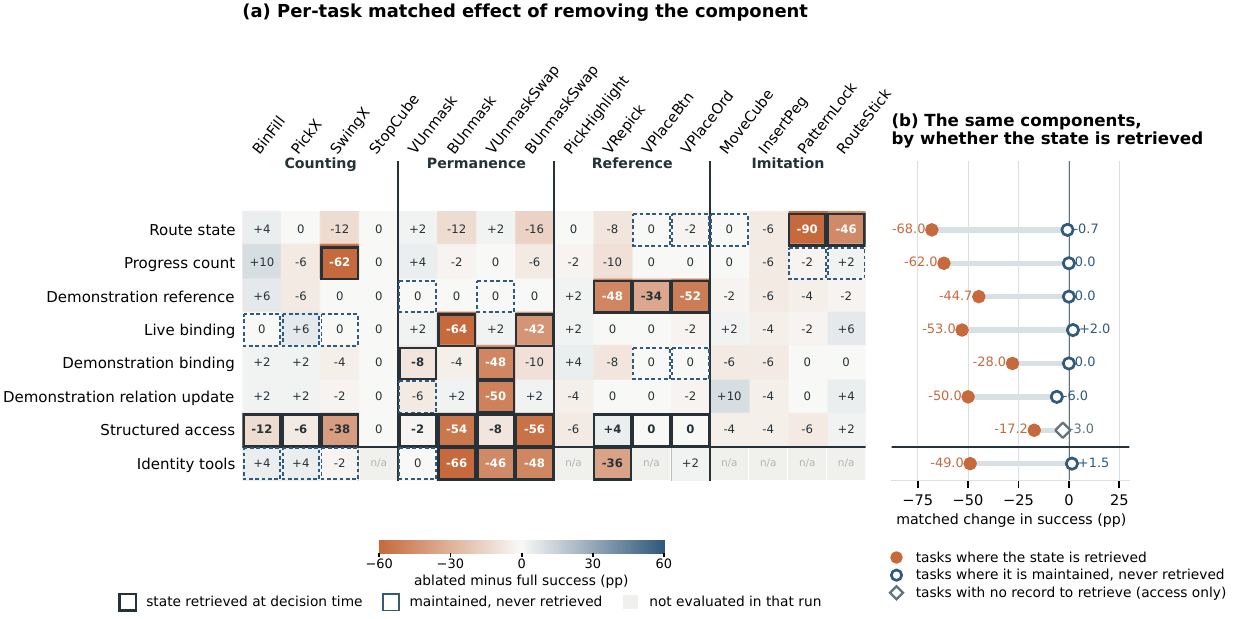}
\caption{\textbf{Matched ablations localize to tasks that retrieve the state.}
(a) Cells show the matched change in success after removing one component (ablated minus full;
50 paired episodes per task). Solid borders mark tasks that retrieve the maintained state for
control; dashed borders mark tasks where the mechanism maintains state in at least 20\% of episodes
but it is never retrieved. The identity-tools row uses a separate 450-episode evaluation over nine
tasks; unevaluated cells are marked \emph{n/a}. (b) Aggregate effects by retrieval status. Losses are
large when state reaches control and small when it is maintained but never retrieved; diamonds denote access
comparisons with no record to retrieve.}
\label{fig:localization}
\end{figure}

\textbf{Different tasks depend on different maintained state.}
Matched ablations show strong task specificity (Figure~\ref{fig:localization}).
Removing route state reduces success by 68.0 points on PatternLock and RouteStick, while
progress-count removal reduces SwingXtimes by 62.0 points. Removing demonstration-reference
state lowers its three corresponding tasks by 44.7 points on average, and live bindings,
demonstration bindings, and relation updates show the same localized pattern. In contrast,
effects on unrelated tasks are small. Thus, different long-horizon tasks depend on different
task-relevant state variables rather than a single generic memory representation. Exact task
sets and aggregate effects are reported in Appendix Table~\ref{tab:final_ablation_matrix}.

\textbf{State matters when it is used for control.}
The same mechanisms often remain active on tasks that never retrieve their state.
On these comparisons, removing route state changes success by only $-0.7$ points,
progress counting by $0.0$, live binding by $+2.0$, and demonstration-reference or
demonstration-binding state by $0.0$ (Figure~\ref{fig:localization}b).
A separate 450-episode ablation gives the same pattern for perception: removing
the identity feature anchor and optical-flow continuity check reduces success by 49.0 points
when the maintained bindings are later retrieved, but changes performance by only $+1.5$ points
on the tasks where the same tools run without retrieval (Figure~\ref{fig:localization}b; per-task
values in Appendix Table~\ref{tab:identity_tools}). Thus, forming or updating state
alone is insufficient; its effect appears when that state is retrieved to resolve a control
decision.

\textbf{Selective retrieval is a separate design problem.}
Replacing structured retrieval with a VLM decision at every proposed subgoal reduces overall
success by 11.9 points: $-17.2$ points on tasks with retrievable historical state versus
$-3.0$ points on the remaining tasks. Free-form retrieval can expose irrelevant historical
state and overwrite subgoals that should instead be resolved from the current observation.
Maintaining useful state and deciding when it should influence control are therefore distinct
problems.

Together, the ablations support two conclusions: different tasks require different maintained state, and that state improves control only when it is selectively retrieved for a decision that depends on it. Additional robustness analyses are reported in
Appendix~\ref{app:robustness}.
\subsection{RQ3: What Challenges Remain Beyond Memory?}
\label{sec:rq3}

\begin{figure}[t]
\centering
\includegraphics[width=\textwidth]{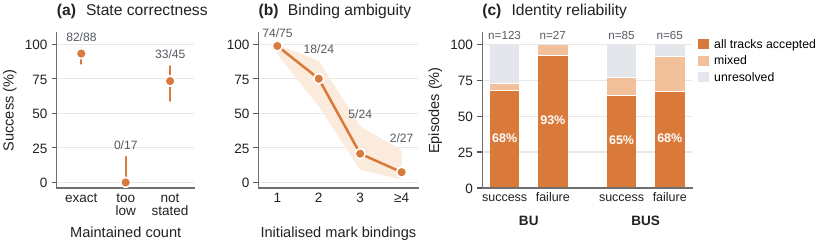}
\caption{\textbf{Representative failure cases.}
(a) SwingXtimes success by count accuracy. (b) PickHighlight success as initialized bindings
increase. (c) Final carrier status for ButtonUnmask (BU) and ButtonUnmaskSwap (BUS). A carrier is a
cover track bound to a remembered cube: ``all tracks accepted'' means every carrier ends internally
verified or reacquired; ``mixed'' combines accepted and uncertain carriers; and ``unresolved'' means
all carriers remain uncertain. These labels are internal states, not ground-truth identity checks.
All panels pool three policy seeds and are diagnostic; bars in (a) and (b) are 95\% Wilson intervals
on the plotted rate.}
\label{fig:rq3_failures}
\end{figure}

Figure~\ref{fig:rq3_failures} examines three representative failure cases in which memory interacts
with state estimation and control. In (a), 82/88 SwingXtimes episodes with an exact count succeed,
versus 0/17 when the count is too low; removing progress state also lowers success by 62 points. In
(b), PickHighlight success falls from 74/75 episodes with one initialized binding to 2/27 with four
or more, showing the difficulty of resolving identities and binding ambiguities. In (c), every carrier
track ends internally accepted (verified or reacquired) in 25/27 ButtonUnmask and 44/65
ButtonUnmaskSwap failures, so an internally coherent track may still contain an incorrect binding or
precede a control error. Other
tasks couple memory with additional capabilities: StopCube benefits from dense local visual history
(Recent4/16/32: 17.3/40.0/41.3\% over three seeds; Appendix~\ref{app:statistics}), while MoveCube and InsertPeg require
fine-grained imitation and execution. RouteStick, by contrast, reaches 53.3\%, near its 55.6\%
oracle-subgoal ceiling. RoboMME
therefore evaluates memory jointly with temporal perception, imitation, and control.

\section{Conclusion and Limitations}
We formulate robot memory as task-conditioned maintenance of policy-sufficient state and introduce
SimpleARM, a training-free layer that specifies candidate state variables, maintains them online,
retrieves them through structured access, and re-grounds recalled entities while leaving the
subgoal predictor and VLA frozen. SimpleARM achieves
\rmmFinalMean\% success on RoboMME, while matched ablations show that relation, reference, progress,
and route state matter primarily where they are retrieved for control. These results support
representing interaction-derived state rather than relying on retained visual history alone.
SimpleARM targets episode-level state maintenance for hierarchical manipulation rather than
high-frequency motion estimation or low-level control. Extending the schema to additional
policy-relevant variables and evaluating it on real robots are important next steps.

\bibliography{references}
\bibliographystyle{iclr2027_conference}

\clearpage
\appendix
\section{Additional Method and Implementation Details}
\label{app:method_details}

\subsection{Instruction-conditioned specification and access}
\label{app:compiler}

The RoboMME implementation queries the instruction with the same frozen Qwen3-VL-4B-Instruct weights
that serve as the subgoal predictor, with the grounded-subgoal adapter disabled and no image in the
prompt, so the memory layer loads no additional model. Four queries run at the start of each episode.
Two produce the state specification $P$: a prompted vote identifying candidate evidence sources and
historical variables that may need to be maintained, and a single query reading the repetition
structure of the instruction, whose count the progress gate later uses. Two more produce the
retrieval specification $A$: one naming how each monitored entity is distinguished, and one emitting
zero or more schema-validated operations for marked entities, previously handled objects,
event-relative progress, or ordered routes. Each of the three voted stages issues the same prompt at
three line widths and takes the majority; identical wording under different line breaking is what
separates stable answers from unstable ones, since decoding is deterministic. A specification is adopted only when at
least two of the three readings call for it, and the reader program additionally requires two
schema-valid readings: the reader compiler returned three valid readings in 1,614 of the 2,400 final
episodes, two in 552, and fell back to no program on the remaining 234.
Language therefore determines what state may be useful and how it may later be queried, but does
not assign stored values or force memory retrieval. Perceptual and event evidence establish state
online, while the structural query in Sec.~\ref{sec:memory_access} determines whether maintained
history is used for the current subgoal. Demonstrations, when present, are handled by the
corresponding perception and procedure parsers rather than by these instruction stages.

Compilation and policy-facing access are separate. A compiled operation is applied only when a
structural test identifies a compatible field in the proposed subgoal. In the final evaluation,
reader programs are applied on PickHighlight, VideoRepick, and VideoPlaceButton; programs emitted
on several other tasks remain inert, and PickXtimes falls back to no program. Entity records are not
injected merely because they exist, while counts, ordinals, and routes use their corresponding field
parsers. This implements the general contract of Eq.~\ref{eq:selective_access}: read memory when a
proposed action field is underdetermined by the current view and determined by interaction history.
The concrete implementation still contains RoboMME-specific lexical choices---token tests on the
composer's phrase, regular expressions for route and ordinal slots, an appearance vocabulary, and
event keywords in demonstration parsers. These instantiate the access interface; they are not the
framework's definition of robot memory.

No language model participates in the remaining operators: the update $\mathcal U$, the retrieval
rule $\mathcal R$, and the verification and re-grounding step $\mathcal V$ are perception tools,
event predicates, and the structural field test, so every language-model call in the method occurs
before the first action is taken.

The plan's coarse \textsc{Need} field is affirmative in all 2,400 final episodes and therefore does
not gate writes in this benchmark. Its \textsc{Watch} phrases and the observable pixel or event
predicates provide the discriminating decisions. This is why we describe the VLM as proposing what
to monitor and the tools as verifying the resulting state.

\subsection{Typed schemas, verification, and state lifecycle}
\label{app:schemas}

Table~\ref{tab:memory_schema} expands the three stores in Eq.~\ref{eq:typed_memory}. All records are
episode-scoped. A candidate write is admitted only after its type-specific observable predicate is
satisfied. Entity records then move through \emph{candidate}, \emph{verified}, \emph{uncertain}, and
\emph{reacquired} states. Historical identity can therefore persist while current geometry is
withheld. A verified read supplies historical disambiguation; entity reads additionally require
current-view localization before modifying the subgoal. If a compatible field, verified state, or
safe current grounding is absent, the memory abstains.

\begin{table}[t]
\caption{Typed memory schemas in the RoboMME instantiation. Language selects fields to monitor;
frozen tools supply and verify their values.}
\label{tab:memory_schema}
\centering
\footnotesize
\setlength{\tabcolsep}{4pt}
\renewcommand{\arraystretch}{1.02}
\begin{tabularx}{\textwidth}{@{}lXXX@{}}
\toprule
state type & persistent field & verified update & policy-facing read \\
\midrule
entity / relation & identity, role, container or current relation & detector proposal plus
appearance/geometric consistency & historical referent; current geometry is re-grounded \\
event / progress & event count, repetition index or event-relative stage & task-observable event
predicate & count, ordinal or completion field \\
trajectory / procedure & ordered route elements and progress index & parsed demonstration segment
or verified route transition & next waypoint or ordered procedural element \\
\bottomrule
\end{tabularx}
\end{table}

\paragraph{State semantics.}
The three stores differ in the temporal structure they preserve.
Entity and relation state retains historical identity and task-relevant relations---for example,
an object's role, container association, or demonstrated identity---rather than the observation
in which that relation was established. Current geometry may become uncertain while the
historical identity or relation remains valid; when needed for control, the corresponding
referent is re-grounded in the current view.

Progress state represents quantities induced by interaction history rather than by any single
observation. For example, a counter can be updated on each newly verified event occurrence as
\begin{equation}
c_t = c_{t-1}
+ \mathbf{1}\!\left[e_t\ \text{is newly verified}\right].
\label{eq:app_progress_update}
\end{equation}
Counts, repetition indices, and event-relative stages therefore summarize verified transitions
accumulated over the episode.

Trajectory and procedural state preserves order-sensitive structure. We represent an ordered
procedure as
\begin{equation}
T=(\tau_1,\ldots,\tau_m),
\qquad
j_t\in\{1,\ldots,m\},
\label{eq:app_trajectory_state}
\end{equation}
where $\tau_i$ denotes a waypoint, route segment, or procedural element and $j_t$ records
current progress when applicable. Order is itself part of the maintained state: the same
observation or waypoint may imply different subsequent behavior at different positions in the
procedure.


\subsection{RoboMME tool instantiation}
\label{app:instantiation}

\paragraph{Entity and relation state.}
Grounding DINO~\citep{liu2023groundingdino} proposes candidate boxes, and a calibrated
named-colour pixel predicate verifies candidates before admission. DINOv2-small
features~\citep{oquab2024dinov2} provide an independent identity anchor. Updates combine identity
and geometric continuity; pyramidal Lucas--Kanade optical
flow~\citep{lucas1981iterative,bouguet2001pyramidal} is consulted when motion continuity is needed,
while SAM2.1-tiny masks~\citep{ravi2025sam2} support ambiguous association and recovery.
A recovery can widen the search at most three times. Agreement restores a record to
\emph{verified}; disagreement preserves historical identity while marking current geometry
\emph{uncertain}. When a retrieved identity is needed for control, current-view detection and
localization provide executable geometry; if reliable grounding cannot be recovered, memory
abstains. Motion and mask tools are invoked only when required by the update or recovery process.

\paragraph{Progress and trajectory state.}
RGB event predicates update repetition and flash counts. Demonstration parsers record handled
objects and event-relative targets. The route parser maintains an ordered sequence of spatial
segments together with its current progress index. This representation contains neither
demonstrated grasp style nor insertion side; tasks requiring these additional capabilities are
discussed in RQ3.

\paragraph{Fixed settings.}
The implementation uses write threshold 0.30, DINO similarity threshold 0.65, recovery budget 3,
maximum episode length 1,300, and action horizon 16. The policy is GroundSG checkpoint 79,999 under
configuration \texttt{mme\_vla\_suite}; the composer is Qwen3-VL-4B-Instruct with grounded-subgoal
adapter checkpoint 1,200. All models, prompts, thresholds, and tools remain fixed during evaluation,
and $M_0$ is reset for every episode.

The added mechanism issues exactly ten language-model calls per episode, in every one of the 2,400
final episodes: three for the state specification, three for entity distinguishability, three for the
reader program, and one for the repetition count. All ten are text-only prompts to the frozen
composer weights with the adapter disabled, and all ten precede the first action, so the method adds
no per-step language-model cost. Perception tools use 2.02 seconds per episode on average: 1.10
seconds for detection, 0.65 for visual features, 0.25 for masks, and 0.03 for optical flow. These
costs exclude the frozen composer and VLA shared with the base system.

\section{Evaluation Protocol}
\label{app:statistics}

Interpreting agent comparisons requires reporting the execution harness, including context
construction, tool interaction, and verification~\citep{zhang2026stop,li2026agent}. We specify the memory
controller and tool configuration in Appendices~\ref{app:schemas} and~\ref{app:instantiation},
alongside the evaluation settings below.

The final benchmark comprises three policy seeds and 50 episodes per task and seed. Table
\ref{tab:seeds} reports every seed-level result together with the standard deviation across the
three seeds; the sixteen-task mean is \rmmFinalMean\% with a seed-to-seed standard deviation of
\rmmFinalSd\ points. Published RoboMME rows are compared at the task and
suite aggregate. Separately, we evaluate the released FrameSamp-Modul checkpoint at seed 7 on the
same 800 episode identities as our method. The two runs share task, scene, and demonstration but
were executed on different hosts, so the paired comparison controls episode difficulty without being
a same-session replication.

Recent$N$+ModuL is RoboMME's FrameSamp+ModuL architecture trained and evaluated with the causal
suffix of the latest $N$ native observations, using 16 history tokens per frame (64, 256, and 512
active history tokens for $N=4$, 16, and 32). Each variant is trained once from the $\pi_{0.5}$ base
with training seed 42 for 80,000 updates, and its checkpoint 79,999 is evaluated with the same three
policy seeds as our method (7, 11, and 23; 16 tasks $\times$ 50 episodes per seed). Its 2,400
episodes therefore match our evaluation protocol, but the three rounds repeat one trained model
rather than three training runs. Table~\ref{tab:recent_seeds} lists every round: Recent32 succeeds
on 254, 254, and 248 of 800 episodes ($31.50\pm0.43$\%), Recent16 on 243, 250, and 244
($30.71\pm0.47$\%), and Recent4 on 222, 230, and 208 ($27.50\pm1.39$\%). The seed-7 rounds and
the seed-11/23 rounds ran on different hosts, so host effects are not separated from seed effects.
On StopCube, Recent4, Recent16, and Recent32 succeed on 26, 60, and 62 of 150 episodes (17.3,
40.0, and 41.3\%); these values are used only as a descriptive diagnostic of recent temporal
context.

\begin{table}[t]
\caption{\textbf{Recent-frame baselines over three policy seeds.} RoboMME's FrameSamp+ModuL
architecture trained and evaluated with the latest 4, 16, or 32 native frames. Each variant is trained
once (seed 42) and its final checkpoint is evaluated with policy seeds 7, 11, and 23, 16 tasks $\times$
50 episodes per seed. Task cells are three-seed mean success (\%); the lower block gives each seed's
successes out of 800 and the mean and sample standard deviation across seeds. The Recent32 column
is the one reported in Table~\ref{tab:main}.}
\label{tab:recent_seeds}
\centering
\footnotesize
\setlength{\tabcolsep}{9pt}
\renewcommand{\arraystretch}{1.02}
\begin{tabular}{@{}lrrr@{}}
\toprule
task & Recent4 & Recent16 & Recent32 \\
\midrule
\multicolumn{4}{@{}l}{\itshape Counting} \\
BinFill & 31.3 & 36.7 & 34.0 \\
PickXtimes & 74.0 & 80.0 & 83.3 \\
SwingXtimes & 83.3 & 78.0 & 80.7 \\
StopCube & 17.3 & 40.0 & 41.3 \\
\addlinespace[2pt]
\multicolumn{4}{@{}l}{\itshape Permanence} \\
VideoUnmask & 28.0 & 29.3 & 19.3 \\
ButtonUnmask & 26.7 & 28.7 & 20.0 \\
VideoUnmaskSwap & 22.0 & 28.0 & 23.3 \\
ButtonUnmaskSwap & 16.0 & 9.3 & 28.0 \\
\addlinespace[2pt]
\multicolumn{4}{@{}l}{\itshape Reference} \\
PickHighlight & 18.7 & 28.7 & 26.7 \\
VideoRepick & 14.0 & 18.7 & 26.7 \\
VideoPlaceButton & 32.0 & 30.0 & 26.0 \\
VideoPlaceOrder & 25.3 & 23.3 & 24.7 \\
\addlinespace[2pt]
\multicolumn{4}{@{}l}{\itshape Imitation} \\
MoveCube & 35.3 & 42.0 & 52.0 \\
InsertPeg & 0.7 & 0.7 & 2.0 \\
PatternLock & 8.7 & 11.3 & 8.7 \\
RouteStick & 6.7 & 6.7 & 7.3 \\
\addlinespace[2pt]
\midrule
seed 7 (of 800) & 222 & 243 & 254 \\
seed 11 (of 800) & 230 & 250 & 254 \\
seed 23 (of 800) & 208 & 244 & 248 \\
\textbf{mean $\pm$ sd (\%)} & $\mathbf{27.50}\pm1.39$ & $\mathbf{30.71}\pm0.47$ & $\mathbf{31.50}\pm0.43$ \\
\bottomrule
\end{tabular}
\end{table}

Each of the ten final-method ablations uses seed 11, covers all sixteen tasks, and matches 50
ablated episodes per task to full-method controls with identical task and episode identifiers. We
report the ablated-minus-control success difference together with the number of matched pairs whose
outcome changes in each direction. Predicted-active sets are fixed from the code path and
control-run activity before aggregation; all remaining tasks are retained as negative controls. Each
ablation is run at a single policy seed, so an effect is characterized by the size of the matched
difference and by whether it is concentrated on the predicted-active tasks, not by a multi-seed
replicate. Where a figure draws an interval, it is a 95\% interval on the plotted quantity---a
percentile interval from 20,000 paired bootstrap resamples for a matched difference, and a Wilson
interval for a single success rate---summarizing the spread of the evaluated episode set rather than
serving as a test.

Structured diagnostics record state transitions, reads, abstentions, recovery events, tool
activity, and episode outcomes. They do not contain the source images or a complete step-aligned
oracle action trace. Consequently, they support analyses of internal state and associations with
outcome, but cannot visually certify every accepted identity or uniquely attribute every residual
failure to grounding versus downstream control.

Two references calibrate how large a pooled difference has to be before it is worth interpreting.
One behaviorally equivalent evaluation pair preserves reader routing on all 2,400 episodes yet flips
11.9\% of individual outcomes, changing the mean by only $-0.42$ points; and cells of the final
ablations that are inactive for a task range from roughly $-3.4$ to $+0.9$ points. Both are
comparable to the \rmmFinalSd-point seed-to-seed standard deviation of the final method. We therefore
read pooled effects of this scale as run-to-run variability unless a localized mechanism and a
matched task signature support them.

\begin{table}[t]
\caption{Per-seed success (\%) for the final method. Each seed contains 50 episodes for each of the
sixteen tasks; the last column reports the sample standard deviation across seeds.}
\label{tab:seeds}
\centering
\footnotesize
\setlength{\tabcolsep}{4.2pt}
\renewcommand{\arraystretch}{0.95}
\begin{tabular}{@{}lrrrrr@{}}
\toprule
task & seed 7 & seed 11 & seed 23 & mean & sd \\
\midrule
BinFill & 72 & 54 & 54 & 60.00 & 10.39 \\
PickXtimes & 88 & 92 & 94 & 91.33 & 3.06 \\
SwingXtimes & 78 & 76 & 76 & 76.67 & 1.15 \\
StopCube & 0 & 0 & 0 & 0.00 & 0.00 \\
VideoUnmask & 100 & 94 & 94 & 96.00 & 3.46 \\
ButtonUnmask & 76 & 90 & 80 & 82.00 & 7.21 \\
VideoUnmaskSwap & 86 & 78 & 76 & 80.00 & 5.29 \\
ButtonUnmaskSwap & 52 & 60 & 58 & 56.67 & 4.16 \\
PickHighlight & 68 & 70 & 60 & 66.00 & 5.29 \\
VideoRepick & 74 & 84 & 76 & 78.00 & 5.29 \\
VideoPlaceButton & 82 & 82 & 82 & 82.00 & 0.00 \\
VideoPlaceOrder & 88 & 88 & 88 & 88.00 & 0.00 \\
MoveCube & 66 & 70 & 66 & 67.33 & 2.31 \\
InsertPeg & 2 & 6 & 4 & 4.00 & 2.00 \\
PatternLock & 92 & 96 & 92 & 93.33 & 2.31 \\
RouteStick & 60 & 48 & 52 & 53.33 & 6.11 \\
\midrule
\textbf{AVG} & \textbf{67.75} & \textbf{68.00} & \textbf{65.75} & \textbf{67.17} & \textbf{1.23} \\
\bottomrule
\end{tabular}
\end{table}

\section{Extended Benchmark Results}
\label{app:extended_results}

\subsection{Task-state annotation}
\label{app:annotation}

Table~\ref{tab:annotation} records the smallest history-dependent variables implied by each task's
instruction and oracle subgoal template. The labels are defined from task semantics rather than
observed performance, and they may overlap. For example, ButtonUnmaskSwap combines a persistent
relation with progress, while VideoPlaceButton combines historical identity with event-relative
order. MoveCube and InsertPeg require manipulation choices not encoded by our route state.

\begin{table}[t]
\caption{Task-state annotation derived from the oracle subgoal templates. Labels may overlap.
$^{\ast}$ marks state expressed only by the instruction, not by the oracle subgoal text.}
\label{tab:annotation}
\centering
\footnotesize
\setlength{\tabcolsep}{4pt}
\renewcommand{\arraystretch}{0.96}
\begin{tabularx}{\textwidth}{@{}ll>{\raggedright\arraybackslash}X@{}}
\toprule
suite & task & history-dependent state \\
\midrule
Counting & BinFill & count/progress \\
& PickXtimes & count/progress \\
& SwingXtimes & count/progress \\
& StopCube$^{\ast}$ & count/progress \\
\addlinespace[1pt]
Permanence & VideoUnmask & entity--location relation \\
& ButtonUnmask & entity--location relation \\
& VideoUnmaskSwap & entity--location relation after swap \\
& ButtonUnmaskSwap & entity--location relation; count/progress \\
\addlinespace[1pt]
Reference & PickHighlight & marked entity; count/progress \\
& VideoRepick & demonstrated entity identity; count/progress \\
& VideoPlaceButton & demonstrated entity identity and event-relative order \\
& VideoPlaceOrder & demonstrated entity identity and ordinal order \\
\addlinespace[1pt]
Imitation & MoveCube & demonstrated manipulation procedure \\
& InsertPeg & demonstrated grasp/insertion procedure \\
& PatternLock & ordered spatial route \\
& RouteStick & ordered spatial route and turn direction \\
\bottomrule
\end{tabularx}
\end{table}

The annotation explains where the system comparison is expected to favor explicit state, but it is
not itself causal evidence. In particular, VideoUnmask and BinFill show large or moderate gains over
FrameSamp despite weak sensitivity to the tested memory ablations; PickHighlight's mark pathway
has no removal ablation. The all-task ablations below supply the causal attribution.

\subsection{Episode-matched FrameSamp comparison}
\label{app:paired_framesamp}

\begin{table}[t]
\caption{\textbf{Episode-matched comparison with the released FrameSamp-Modul checkpoint.}
The runs share task, scene, demonstration, and episode identity but were executed on different
hosts. Tasks are grouped into the four RoboMME suites; both halves of the table have the same
columns. \emph{FS} abbreviates FrameSamp-Modul, $\Delta$ is ours minus FrameSamp in percentage
points, and \emph{only} counts the episodes solved by exactly one system, as
ours-only\,/\,FrameSamp-only.}
\label{tab:framesamp_paired}
\centering
\scriptsize
\setlength{\tabcolsep}{3pt}
\renewcommand{\arraystretch}{1.06}
\begin{tabular}{@{}lrrrr@{\hspace{1.5em}}lrrrr@{}}
\toprule
task & ours & FS & $\Delta$ & only & task & ours & FS & $\Delta$ & only \\
\midrule
\multicolumn{5}{@{}l}{\itshape Counting} & \multicolumn{5}{@{}l}{\itshape Reference} \\
BinFill & 36/50 & 22/50 & $+28$ & 17/3 & PickHighlight & 34/50 & 9/50 & $+50$ & 26/1 \\
PickXtimes & 44/50 & 48/50 & $-8$ & 1/5 & VideoRepick & 37/50 & 17/50 & $+40$ & 22/2 \\
SwingXtimes & 39/50 & 44/50 & $-10$ & 4/9 & VideoPlaceButton & 41/50 & 28/50 & $+26$ & 15/2 \\
StopCube & 0/50 & 25/50 & $-50$ & 0/25 & VideoPlaceOrder & 44/50 & 22/50 & $+44$ & 23/1 \\
\cmidrule(r{1.5em}){1-5}\cmidrule{6-10}
\textit{suite} & 119/200 & 139/200 & $-10.0$ & 22/42 & \textit{suite} & 156/200 & 76/200 & $\mathbf{+40.0}$ & 86/6 \\
\addlinespace[3pt]
\multicolumn{5}{@{}l}{\itshape Permanence} & \multicolumn{5}{@{}l}{\itshape Imitation} \\
VideoUnmask & 50/50 & 17/50 & $+66$ & 33/0 & MoveCube & 33/50 & 41/50 & $-16$ & 6/14 \\
ButtonUnmask & 38/50 & 10/50 & $+56$ & 28/0 & InsertPeg & 1/50 & 5/50 & $-8$ & 1/5 \\
VideoUnmaskSwap & 43/50 & 13/50 & $+60$ & 32/2 & PatternLock & 46/50 & 26/50 & $+40$ & 20/0 \\
ButtonUnmaskSwap & 26/50 & 8/50 & $+36$ & 19/1 & RouteStick & 30/50 & 34/50 & $-8$ & 7/11 \\
\cmidrule(r{1.5em}){1-5}\cmidrule{6-10}
\textit{suite} & 157/200 & 48/200 & $\mathbf{+54.5}$ & 112/3 & \textit{suite} & 110/200 & 106/200 & $+2.0$ & 34/30 \\
\midrule
\multicolumn{10}{@{}l}{\textbf{All 16 tasks:}\quad ours \textbf{542/800}\quad FS \textbf{369/800}%
\quad $\Delta=\mathbf{+21.6}$\quad only \textbf{254/81}} \\
\bottomrule
\end{tabular}
\end{table}

Table~\ref{tab:framesamp_paired} lists every task. The reproduced checkpoint reaches 46.12\%, close to the published three-seed 44.51\%. The paired
difference is 21.6 points, with 254 ours-only and 81 FrameSamp-only successes. Ten of the sixteen
tasks favor our method by at least 26 points; StopCube is the only large loss. The suite rows locate
the gain: $+54.5$ points on Permanence and $+40.0$ on Reference, against $+2.0$ on Imitation and
$-10.0$ on Counting. Because the runs
share episode identities but not host-level execution noise, this is a paired comparison of the two
systems at the task level, not a same-session replication.

\section{Full Mechanism Ablations}
\label{app:ablations}

\begin{table}[t]
\caption{All final-method ablations on the common 800-episode seed-11 pool. Each row compares
the ablation with its episode-matched full-method control, ordered by the size of the effect.
\emph{better/worse} counts the matched pairs whose outcome changes in each direction, so their sum
is the number of episodes the ablation altered at all. Row order matches
Figure~\ref{fig:ablation_forest}.}
\label{tab:final_ablations_full}
\centering
\small
\setlength{\tabcolsep}{8pt}
\renewcommand{\arraystretch}{1.08}
\begin{tabular}{@{}lrrrr@{}}
\toprule
ablation & control & ablated & $\Delta$ & better/worse \\
\midrule
\multicolumn{5}{@{}l}{\itshape Memory components} \\
agent-decided access & 67.38 & 55.50 & $\mathbf{-11.88}$ & 28/123 \\
no route state & 67.75 & 56.25 & $\mathbf{-11.50}$ & 24/116 \\
no demo references & 67.25 & 58.12 & $\mathbf{-9.12}$ & 43/116 \\
no live identity track & 67.50 & 61.62 & $-5.88$ & 42/89 \\
no demo notes & 67.75 & 62.38 & $-5.38$ & 38/81 \\
no flash-count state & 67.62 & 62.62 & $-5.00$ & 40/80 \\
no demo following & 67.75 & 64.88 & $-2.88$ & 42/65 \\
\addlinespace[3pt]
\multicolumn{5}{@{}l}{\itshape Robustness and implementation controls} \\
single-vote write plan & 67.62 & 64.62 & $-3.00$ & 39/63 \\
refine every verb & 67.38 & 65.38 & $-2.00$ & 47/63 \\
source from environment & 67.25 & 67.25 & $0.00$ & 44/44 \\
\bottomrule
\end{tabular}
\end{table}

\label{app:ablation_details}


\begin{table}[t]
\caption{\textbf{Task specificity of memory-component ablations.}
Each ablation is evaluated on all 16 tasks with 50 matched episodes per task.
Task sets are specified \emph{a priori} from the state variable or access mechanism
required by the final method. $\Delta$ denotes the change in success rate relative
to the full method, in percentage points.}
\label{tab:final_ablation_matrix}

\centering
\small
\setlength{\tabcolsep}{4pt}
\renewcommand{\arraystretch}{1.10}

\begin{tabularx}{\textwidth}{
@{}
>{\raggedright\arraybackslash}p{1.55cm}
>{\raggedright\arraybackslash}p{2.55cm}
>{\raggedright\arraybackslash}X
>{\raggedleft\arraybackslash}p{1.35cm}
>{\raggedleft\arraybackslash}p{1.45cm}
@{}}
\toprule
\textbf{Category} &
\textbf{Ablation} &
\textbf{Tasks requiring component} &
\textbf{$\Delta$ required} &
\textbf{$\Delta$ remaining} \\
\midrule

\multirow{5}{1.55cm}{\raggedright\textit{Maintained state}}
& Route-state removal
& PatternLock, RouteStick
& $\mathbf{-68.0}$
& $-3.4$ \\

& Progress-count removal
& SwingXtimes
& $\mathbf{-62.0}$
& $-1.2$ \\

& Demonstration-reference removal
& VideoRepick, VideoPlaceButton, VideoPlaceOrder
& $\mathbf{-44.7}$
& $-0.9$ \\

& Live-binding removal
& ButtonUnmask, ButtonUnmaskSwap
& $\mathbf{-53.0}$
& $+0.9$ \\

& Demonstration-binding removal
& VideoUnmask, VideoUnmaskSwap
& $\mathbf{-28.0}$
& $-2.1$ \\

\cmidrule(l){2-5}

\textit{State update}
& Demonstration relation update
& VideoUnmaskSwap
& $\mathbf{-50.0}$
& $+0.3$ \\

\cmidrule(l){2-5}

\textit{Memory access}
& Agent-decided access
& Ten note-bearing tasks
& $\mathbf{-17.2}$
& $-3.0$ \\

\bottomrule
\end{tabularx}
\end{table}

\begin{figure}[t]
\centering
\includegraphics[width=0.98\textwidth]{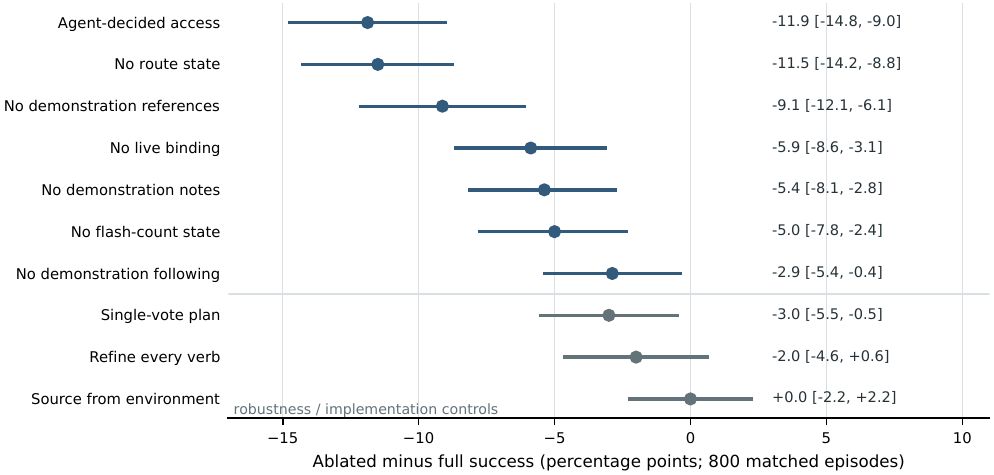}
\caption{All ten final-method ablations on 800 episode-matched comparisons each. Points are
ablated-minus-full success and lines are paired-bootstrap 95\% intervals. The lower three rows are
robustness controls rather than typed memory mechanisms.}
\label{fig:ablation_forest}
\end{figure}

All ten ablations in Table~\ref{tab:final_ablations_full} cover the same sixteen tasks and 800
matched episodes. Figure~\ref{fig:ablation_forest} shows the pooled effects; Table
\ref{tab:ablation_by_task} gives every task cell. The pooled score is useful for overall impact, but
the active/off-target split in Table~\ref{tab:final_ablation_matrix} is the direct test of mechanism
specificity.
Table~\ref{tab:final_ablation_matrix} reports the task sets and aggregate matched
effects used in the RQ2 analysis. Task sets are fixed from the state variable or
access mechanism exercised by the final method; all remaining tasks serve as
negative controls.

\begin{table}[t]
\caption{Complete ablation--task matrix. Every cell contains 50 matched episodes at seed 11;
entries are ablated minus full success (percentage points). Bold cells are predicted active before
observing outcomes. Task abbreviations follow their column order in Table~\ref{tab:main}.}
\label{tab:ablation_by_task}
\centering
\scriptsize
\setlength{\tabcolsep}{1.5pt}
\renewcommand{\arraystretch}{0.96}
\resizebox{\textwidth}{!}{%
\begin{tabular}{@{}lrrrrrrrrrrrrrrrr@{}}
\toprule
& BF & PX & SX & SC & VU & BU & VUS & BUS & PH & VR & VPB & VPO & MC & IP & PL & RS \\
\midrule
no route & $+4$ & $0$ & $-12$ & $0$ & $+2$ & $-12$ & $+2$ & $-16$ & $0$ & $-8$ & $0$ & $-2$ & $0$ & $-6$ & $\mathbf{-90}$ & $\mathbf{-46}$ \\
no flash count & $+10$ & $-6$ & $\mathbf{-62}$ & $0$ & $+4$ & $-2$ & $0$ & $-6$ & $-2$ & $-10$ & $0$ & $0$ & $0$ & $-6$ & $-2$ & $+2$ \\
no demo references & $+6$ & $-6$ & $0$ & $0$ & $0$ & $0$ & $0$ & $0$ & $+2$ & $\mathbf{-48}$ & $\mathbf{-34}$ & $\mathbf{-52}$ & $-2$ & $-6$ & $-4$ & $-2$ \\
no live track & $0$ & $+6$ & $0$ & $0$ & $+2$ & $\mathbf{-64}$ & $+2$ & $\mathbf{-42}$ & $+2$ & $0$ & $0$ & $-2$ & $+2$ & $-4$ & $-2$ & $+6$ \\
no demo notes & $+2$ & $+2$ & $-4$ & $0$ & $\mathbf{-8}$ & $-4$ & $\mathbf{-48}$ & $-10$ & $+4$ & $-8$ & $0$ & $0$ & $-6$ & $-6$ & $0$ & $0$ \\
no demo following & $+2$ & $+2$ & $-2$ & $0$ & $-6$ & $+2$ & $\mathbf{-50}$ & $+2$ & $-4$ & $0$ & $0$ & $-2$ & $+10$ & $-4$ & $0$ & $+4$ \\
agent access & $\mathbf{-12}$ & $\mathbf{-6}$ & $\mathbf{-38}$ & $0$ & $\mathbf{-2}$ & $\mathbf{-54}$ & $\mathbf{-8}$ & $\mathbf{-56}$ & $-6$ & $\mathbf{+4}$ & $\mathbf{0}$ & $\mathbf{0}$ & $-4$ & $-4$ & $-6$ & $+2$ \\
\midrule
single vote & $-4$ & $-2$ & $-8$ & $0$ & $-4$ & $-10$ & $-8$ & $-12$ & $+2$ & $-4$ & $0$ & $-2$ & $+6$ & $0$ & $-2$ & $0$ \\
refine all verbs & $+6$ & $-2$ & $+4$ & $0$ & $0$ & $-20$ & $-4$ & $-22$ & $+4$ & $-2$ & $0$ & $0$ & $+6$ & $-4$ & $-8$ & $+10$ \\
source = environment & $+8$ & $-2$ & $-6$ & $0$ & $+2$ & $-4$ & $+2$ & $-2$ & $-6$ & $0$ & $0$ & $-2$ & $+8$ & $-2$ & $0$ & $+4$ \\
\bottomrule
\end{tabular}}
\end{table}

The matrix reveals both necessity and negative-control specificity. PatternLock and RouteStick lose
90 and 46 points without route state; SwingXtimes loses 62 without the flash count; the three
demonstration-reference tasks lose 34--52 without reference state; the ButtonUnmask pair loses 64
and 42 without the live binding; and VideoUnmaskSwap loses 50 without demonstration cover-following.
Off-target changes are much smaller. The largest of them---no route on ButtonUnmaskSwap
($-16$ points)---occurs on a task where route state is never formed, so it cannot be an effect of
removing that state; single cells of this size are read against the run-to-run variability band in
Appendix~\ref{app:statistics} rather than on their own.

\subsection{Activity, access, and compiled readers}
\label{app:activity}

\begin{table}[t]
\caption{Mechanism activity does not imply contribution to control. Activity is the fraction of
final-run episodes in which the mechanism is exercised; sensitive tasks are those affected by the
corresponding ablation.}
\label{tab:mechanism_activity}
\centering
\scriptsize
\setlength{\tabcolsep}{3pt}
\renewcommand{\arraystretch}{0.98}
\begin{tabularx}{\textwidth}{@{}p{2.3cm}>{\raggedright\arraybackslash}X>{\raggedright\arraybackslash}p{3.0cm}@{}}
\toprule
mechanism & active tasks (fraction of episodes) & ablation-sensitive tasks \\
\midrule
live carrier track & BinFill 44\%, PickXtimes 71\%, SwingXtimes 77\%, ButtonUnmask 100\%, ButtonUnmaskSwap 100\% & ButtonUnmask, ButtonUnmaskSwap \\
demo cover track & VideoUnmask 100\%, VideoUnmaskSwap 100\% & VideoUnmaskSwap \\
flash counter & SwingXtimes 96\%, PatternLock 47\%, RouteStick 47\% & SwingXtimes \\
route fill & PatternLock 100\%, RouteStick 100\% & PatternLock, RouteStick \\
entity read & VideoUnmask 96\%, ButtonUnmask 100\%, VideoUnmaskSwap 92\%, ButtonUnmaskSwap 100\%, VideoRepick 100\% & ButtonUnmask, ButtonUnmaskSwap \\
mark reader & PickHighlight 100\% & not tested \\
SAM & seven tasks, 54--79\% & not tested \\
\bottomrule
\end{tabularx}
\end{table}

Table~\ref{tab:mechanism_activity} lists how often each mechanism runs. Live carrier bindings are formed in 44--77\% of Counting episodes, and the flash counter observes red cues
on route tasks, yet removing those mechanisms has no corresponding effect there. Similarly,
demonstration notes are written in 92--96\% of VideoPlaceButton and VideoPlaceOrder episodes but are
never read, and their removal changes neither task. Activity therefore confirms that a mechanism is
exercised, but does not establish a contribution to control.

Structured access prevents active but irrelevant state from reaching control. Under the structural
rule, entity reads per episode are 0.0 on BinFill, PickXtimes, and SwingXtimes; under agent-decided
access they rise to 37.2, 35.5, and 29.0. On ButtonUnmask and ButtonUnmaskSwap they rise from 12.9 to
34.0 and from 16.9 to 43.0. In worsened SwingXtimes pairs, the free-form reader exposes a plausible
carrier track to pick and move subgoals, replacing the current-view coordinate and disrupting the
subsequent count. This explains why structured access is a mechanism rather than merely a cheaper
retrieval policy.

The reader compiler is deliberately separated from this gate. Its programs are correctly applied
on PickHighlight (mark), VideoRepick (handled object), and VideoPlaceButton (sequence). It also emits
inert programs on seven live or non-reference tasks, falls back on all PickXtimes episodes, and falls
back on half of BinFill. These outputs do not affect control unless the structural gate finds a
compatible history-dependent field. The evidence therefore supports agentic candidate specification
plus precision-oriented access, not a claim of perfect compiler routing.

Table~\ref{tab:identity_tools} reports the per-task values underlying the identity-tools comparison
in Figure~\ref{fig:localization}b. Unlike the typed-state removals, this ablation is exercised
across both groups: it produces large losses when the maintained binding is retrieved and almost no
change when the same tools run without the binding reaching control.

\begin{table}[t]
\caption{\textbf{Matched identity-tool ablation.} The frozen feature anchor and
optical-flow continuity check are removed while the color verifier is retained.
The comparison covers nine tasks with 50 matched episodes each. ``Retrieved'' means
that the maintained binding is read at decision time and conditions the action; the
pooled ``maintained, not retrieved'' row is the open circle of Figure~\ref{fig:localization}b
(the four tasks on which the tools run in at least 20\% of episodes), and ``all other
tasks'' adds VideoPlaceOrder, where they run in 8\%. SAM is absent
from both arms, so this comparison should not be interpreted as removing every
identity tool.}
\label{tab:identity_tools}
\centering
\small
\setlength{\tabcolsep}{4pt}
\renewcommand{\arraystretch}{1.02}
\begin{tabular}{@{}lccrr@{}}
\toprule
Task & Binding retrieved & Full & Ablated & $\Delta$ \\
\midrule
ButtonUnmask & yes & 42/50 & 9/50 & $\mathbf{-66}$ \\
ButtonUnmaskSwap & yes & 34/50 & 10/50 & $\mathbf{-48}$ \\
VideoUnmaskSwap & yes & 37/50 & 14/50 & $\mathbf{-46}$ \\
VideoRepick & yes & 34/50 & 16/50 & $\mathbf{-36}$ \\
\addlinespace[1pt]
VideoUnmask & no & 48/50 & 48/50 & $0$ \\
SwingXtimes & no & 34/50 & 33/50 & $-2$ \\
PickXtimes & no & 46/50 & 48/50 & $+4$ \\
BinFill & no & 26/50 & 28/50 & $+4$ \\
VideoPlaceOrder & no & 42/50 & 43/50 & $+2$ \\
\midrule
Retrieved tasks & & 147/200 & 49/200 & $\mathbf{-49.0}$ \\
Maintained, not retrieved (4 tasks) & & 154/200 & 157/200 & $+1.5$ \\
All other tasks (5 tasks) & & 196/250 & 200/250 & $+1.6$ \\
\bottomrule
\end{tabular}
\end{table}

\subsection{Robustness controls}
\label{app:robustness}

Forcing one planning vote changes the pooled score by $-3.0$ points, refining every verb by $-2.0$,
and forcing the evidence source to environment by $0.0$. The first two pooled differences are
comparable with the between-session drift band and are not treated as core mechanisms. The
refinement control nevertheless has a localized timing effect: ButtonUnmask and
ButtonUnmaskSwap lose 20 and 22 points, because refining button-press coordinates increases composer
calls and delays the prerequisite before the historical container read. This is evidence about
downstream execution, not about memory content.

\begin{table}[t]
\caption{\textbf{Supplementary robustness comparisons.} Each row is matched by
task and episode to a configuration-matched control. These comparisons use a
distinct evaluation configuration and therefore characterize sensitivity rather
than final-method component necessity. $\Delta$ is variant minus control in
success percentage points.}
\label{tab:supp_robustness}
\centering
\small
\setlength{\tabcolsep}{4pt}
\renewcommand{\arraystretch}{1.03}
\begin{tabularx}{\textwidth}{@{}>{\raggedright\arraybackslash}Xrrr@{}}
\toprule
Comparison & Tasks & Matched episodes & $\Delta$ \\
\midrule
Instruction rewording & 8 & 416 & $-17.8$ \\
\quad with environment-provided evidence source & 8 & 416 & $-7.7$ \\
Detector substitution & 8 & 165 & $-5.5$ \\
Keyword rather than agent-specified write policy & 11 & 422 & $-4.5$ \\
Write gate removed & 11 & 548 & $-5.1$ \\
Structural access replaced by loose lexical access & 6 & 222 & $-5.9$ \\
\bottomrule
\end{tabularx}
\end{table}

Table~\ref{tab:supp_robustness} separates robustness analyses from the final component ablations.
Instruction rewording produces the largest sensitivity; providing the evidence source from the
environment reduces, but does not eliminate, that loss. Detector substitution, write-policy changes,
removing the write gate, and loosening the structural access test have smaller effects. Because these
comparisons use a distinct evaluation configuration, we do not combine them with the matched
final-method ablations in Figure~\ref{fig:localization} or Table~\ref{tab:final_ablation_matrix}.

\subsection{Optical-flow control under a different tool stack}
\label{app:flow_ablation}

\begin{table}[t]
\caption{Matched seed-7 no-flow control on five identity/update tasks. Both arms ran on a host where
SAM was unavailable, unlike the final method; therefore this does not estimate flow's marginal
effect in the shipped tool stack. $\Delta$ is no-flow minus control, and \emph{better/worse}
counts the matched episodes whose outcome changes in each direction.}
\label{tab:flow_ablation}
\centering
\small
\setlength{\tabcolsep}{5pt}
\renewcommand{\arraystretch}{1.02}
\begin{tabularx}{\textwidth}{@{}l>{\raggedright\arraybackslash}Xccc@{}}
\toprule
scope & update condition & control $\rightarrow$ no flow & $\Delta$ & better/worse \\
\midrule
ButtonUnmaskSwap & bound cover moves after occlusion & $34/50\rightarrow22/50$ & $-24$ & 2/14 \\
other four tasks & no live post-cover update of this type & $161/200\rightarrow161/200$ & $0$ & 12/12 \\
\midrule
\textbf{pooled} & five predefined tasks & $\mathbf{195/250\rightarrow183/250}$ & $\mathbf{-4.8}$ & 14/26 \\
\bottomrule
\end{tabularx}
\end{table}

The no-flow comparison (Table~\ref{tab:flow_ablation}) is not part of the final ten-ablation set. Both its ablated and control arms
ran on a host where SAM could not be imported, whereas SAM is available in the final method and is
used for ambiguous association and reacquisition. The localized ButtonUnmaskSwap loss is therefore
evidence that flow can support motion continuity when no mask fallback is available; it does not
establish that flow is necessary in the final evaluated configuration. A matched no-flow run with SAM available
would be required for that claim.

\section{Failure Diagnostics and State Coverage}
\label{app:failure_analysis}

\subsection{Representational coverage}

Table~\ref{tab:schema_coverage} summarizes representative tasks for which the current schema does
or does not contain the required control variable. It interprets task requirements together with
the matched ablations and oracle headroom; it is not an additional performance comparison.

\begin{table}[t]
\caption{\textbf{Representational coverage of the current memory schema.} Matched removals quantify
the contribution of represented variables; uncovered tasks require a control variable absent from
the schema. ``Represented'' does not imply that the complete controller succeeds.}
\label{tab:schema_coverage}
\centering
\scriptsize
\setlength{\tabcolsep}{3.2pt}
\renewcommand{\arraystretch}{0.98}
\begin{tabularx}{\textwidth}{@{}lp{2.65cm}p{1.55cm}>{\raggedright\arraybackslash}X@{}}
\toprule
task & required policy variable & represented? & evidence \\
\midrule
SwingXtimes & accumulated swing count & \multirow{5}{*}{yes} & no count: $-62$ points \\
ButtonUnmask & live cube--container binding & & no binding: $-64$ points \\
VideoPlaceOrder & demonstrated ordinal target & & no reference: $-52$ points \\
PatternLock & ordered route & & no route: $-90$ points \\
RouteStick & ordered route and turn sense & & 53.3\%, near 55.6\% oracle ceiling \\
\midrule
StopCube & timed reach count of moving cube & \multirow{3}{*}{no} & 0/150; no state formed \\
MoveCube & demonstrated manipulation manner & & no tested ablation changes outcome materially \\
InsertPeg & demonstrated peg end and insertion side & & no corresponding memory field \\
\bottomrule
\end{tabularx}
\end{table}

\subsection{Memory-state attainment across tasks}

\begin{table}[t]
\caption{Memory-state attainment among the 788 failed episodes. Each count records the furthest
stage observed at any point in the episode, not the stage that caused failure. Specialized
reference access in VideoPlaceButton and VideoPlaceOrder is not represented by the same access
statistic and is therefore reported separately.}
\label{tab:memory_attainment}
\centering
\scriptsize
\setlength{\tabcolsep}{3.2pt}
\renewcommand{\arraystretch}{0.94}
\begin{tabularx}{\textwidth}{@{}lr>{\raggedright\arraybackslash}X>{\raggedright\arraybackslash}X@{}}
\toprule
task & failures & furthest recorded attainment (episodes) & task-level interpretation \\
\midrule
BinFill & 60 & state formed, no recorded access: 60 & count state maintained \\
PickXtimes & 13 & state formed, no recorded access: 13 & count state maintained \\
SwingXtimes & 35 & state formed; ordinal access analyzed separately & progress state maintained \\
StopCube & 150 & no usable typed state: 150 & required timed state absent \\
VideoUnmask & 6 & state returned: 5; formed without access: 1 & demonstration relation \\
ButtonUnmask & 27 & state returned: 27 & live entity--carrier binding \\
VideoUnmaskSwap & 30 & state returned: 21; all reads abstained: 3; no access: 6 & moving demonstration relation \\
ButtonUnmaskSwap & 65 & state returned: 65 & live moving binding \\
PickHighlight & 51 & mark relation returned: 51 & historical mark binding \\
VideoRepick & 33 & reference returned: 33 & demonstrated identity \\
VideoPlaceButton & 27 & specialized reference path; common access statistic unavailable & demonstrated target \\
VideoPlaceOrder & 18 & specialized reference path; common access statistic unavailable & demonstrated ordinal target \\
MoveCube & 49 & no usable state: 40; formed without access: 9 & demonstrated manner absent \\
InsertPeg & 144 & no usable state: 132; formed without access: 12 & peg end/side absent \\
PatternLock & 10 & route returned: 10 & ordered route \\
RouteStick & 70 & route returned: 70 & ordered route \\
\bottomrule
\end{tabularx}
\end{table}

Table~\ref{tab:memory_attainment} summarizes how far failed episodes engage the available memory
states. The categories are computed from episode-level totals and therefore report the furthest
attained state or access event, not when or why failure occurs. In 12 of 16 tasks, successful and
failed episodes receive essentially the same attainment label; 559/788 failures occur on tasks for
which that label is constant. The table is therefore a coverage description rather than a failure
localization. VideoPlaceButton and VideoPlaceOrder use a specialized demonstration-reference path
that is not represented by the common entity/event-access statistic, so their access is not assigned
to a directly comparable category. Matched ablations in RQ2, rather than these observational
counts, provide causal evidence.

\subsection{State correctness and binding complexity}

For SwingXtimes goals that specify two or three repetitions, all 17 episodes with an under-counted
state fail, while 82/88 with the exact count succeed. Combined with the matched no-count
ablation, this links missed event detections to failure without claiming that an exact count is
sufficient. For PickHighlight, initialized bindings of one, two, three, and at least four yield
74/75, 18/24, 5/24, and 2/27 successes. The trend is diagnostic because no matched ablation
removes the mark-memory pathway.

On ButtonUnmask, failures average 7.19 rejected tracks and 17.48 exhausted recoveries per episode,
versus 1.23 and 2.63 for successes. On ButtonUnmaskSwap the corresponding averages are 12.71 and
40.22 versus 2.92 and 4.24. Only 6\% and 10\% of failed-episode read attempts abstain, and every
carrier ends internally accepted in 25/27 and 44/65 failures (verified in 22/27 and 39/65, the rest
reacquired; Figure~\ref{fig:rq3_failures}c). Internal verification reflects
the method's identity and geometry checks, not ground-truth identity. The pattern is consistent with
identity errors that remain undetected or with later control failure; the available diagnostics do
not separate these possibilities.

\subsection{Route length and downstream execution}

\begin{table}[t]
\caption{Final-method success by parsed route length. These are observational difficulty slices,
not controlled ablations; small non-monotonic cells, especially length seven, should not be overread.}
\label{tab:route_depth}
\centering
\small
\setlength{\tabcolsep}{5pt}
\begin{tabular}{@{}crrcrr@{}}
\toprule
& \multicolumn{2}{c}{PatternLock} && \multicolumn{2}{c}{RouteStick} \\
\cmidrule(lr){2-3}\cmidrule(lr){5-6}
route length & $n$ & success & & $n$ & success \\
\midrule
1 & 24 & 100.0\% && --- & --- \\
2 & 27 & 100.0\% && 42 & 76.2\% \\
3 & 54 & 100.0\% && 36 & 58.3\% \\
4 & 24 & 91.7\% && 27 & 40.7\% \\
5 & 3 & 33.3\% && 27 & 25.9\% \\
6 & 12 & 75.0\% && 9 & 11.1\% \\
7 & 6 & 50.0\% && 9 & 88.9\% \\
\bottomrule
\end{tabular}
\end{table}

Table~\ref{tab:route_depth} slices the two route tasks by parsed route length. PatternLock remains reliable through route length three and then becomes less stable. RouteStick
declines broadly from lengths two through six, although the small length-seven cell is
non-monotonic. Route length is not randomized and can correlate with geometry, so these are
difficulty slices rather than causal length effects. None of the 10 PatternLock or 70 RouteStick
failures reaches the 80-subgoal interaction horizon. By contrast, MoveCube reaches that horizon in
37/49 failures (75.5\%) and InsertPeg in 61/144 (42.4\%), indicating missing schema fields together
with difficult downstream execution.

\subsection{Seed-invariant demonstration failures}

VideoRepick's internal \texttt{correct cube} record is marked verified in all 150 episodes,
including all 33 failures. VideoPlaceButton has the same 41 successful episode identities at all
three policy seeds; VideoPlaceOrder shares 43 of 44 successful identities. These outcomes suggest
that residual demonstration-reference failures are fixed primarily by how the demonstration is
interpreted, rather than by policy-seed variation. Internal verification still does not certify
that the selected demonstration entity is correct.

\subsection{Episode-level state-to-action lifecycle}
\label{app:qualitative_lifecycle}

\begin{figure}[p]
\centering
\includegraphics[width=\textwidth]{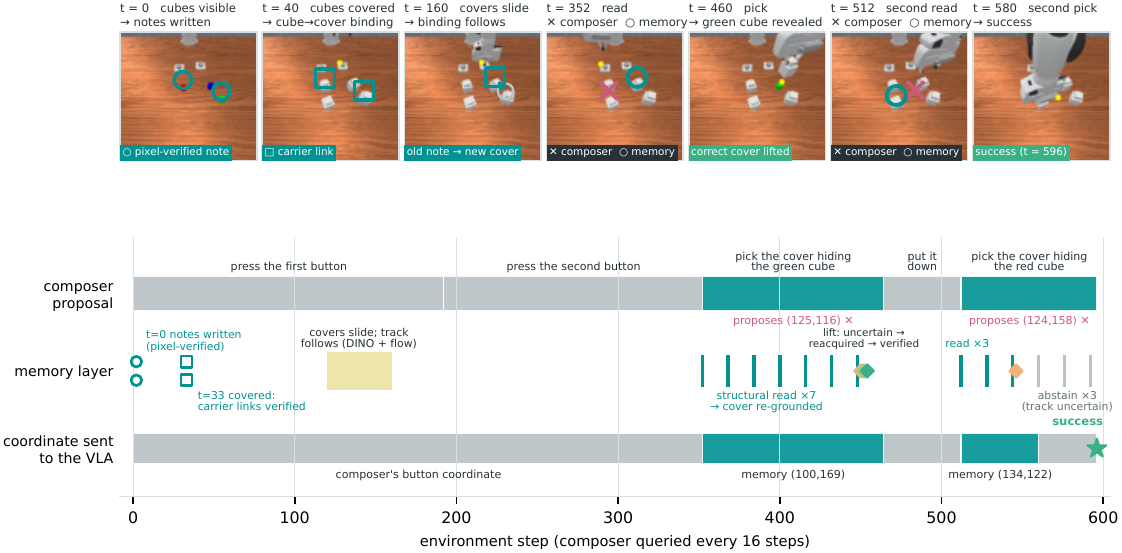}
\caption{\textbf{A complete ButtonUnmaskSwap memory lifecycle (seed 7, episode 47).} The memory
writes cube identities while they are visible, converts them to cube--cover bindings after
occlusion, and updates the bindings as the covers move. At both history-dependent reads, the frozen
composer proposes the wrong container; the memory retrieves the corresponding binding, re-grounds
it to current geometry, and sends the corrected coordinate to the frozen VLA. The bottom lanes show
the composer proposal, memory operations, and coordinate used for control.}
\label{fig:method_episode}
\end{figure}

Figure~\ref{fig:method_episode} follows a complete ButtonUnmaskSwap episode. Unlike the aggregate
results, this example shows when state is written, updated, accessed, and re-grounded before it is
passed to the VLA. It is qualitative evidence of mechanism operation, not an estimate of accuracy
or necessity.

\subsection{Tool-use examples}
\label{app:tool_examples}

\begin{figure}[p]
\centering
\includegraphics[width=\textwidth]{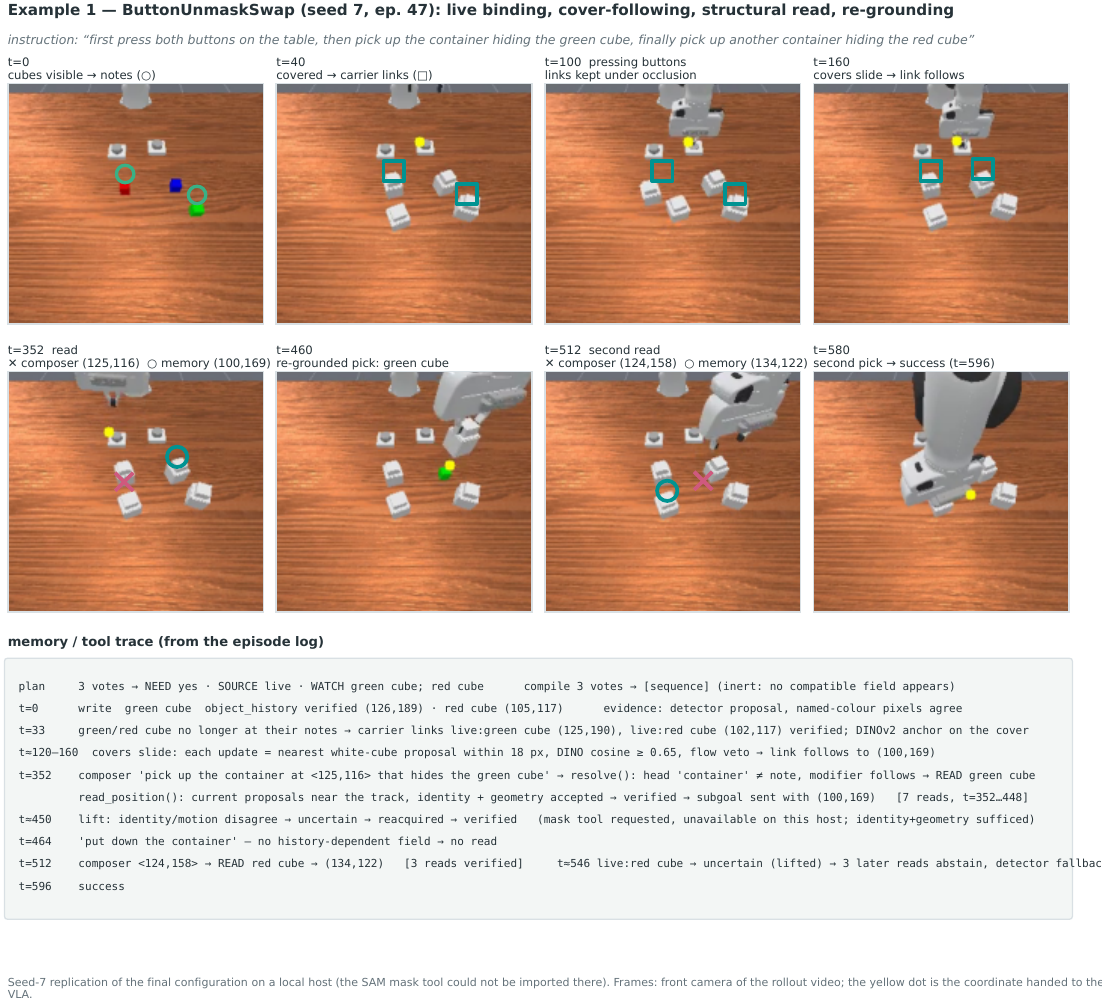}
\caption{Live entity binding, cover following, structured access, and current-view re-grounding on
ButtonUnmaskSwap.}
\label{fig:app_ex1}
\end{figure}

\begin{figure}[p]
\centering
\includegraphics[width=\textwidth]{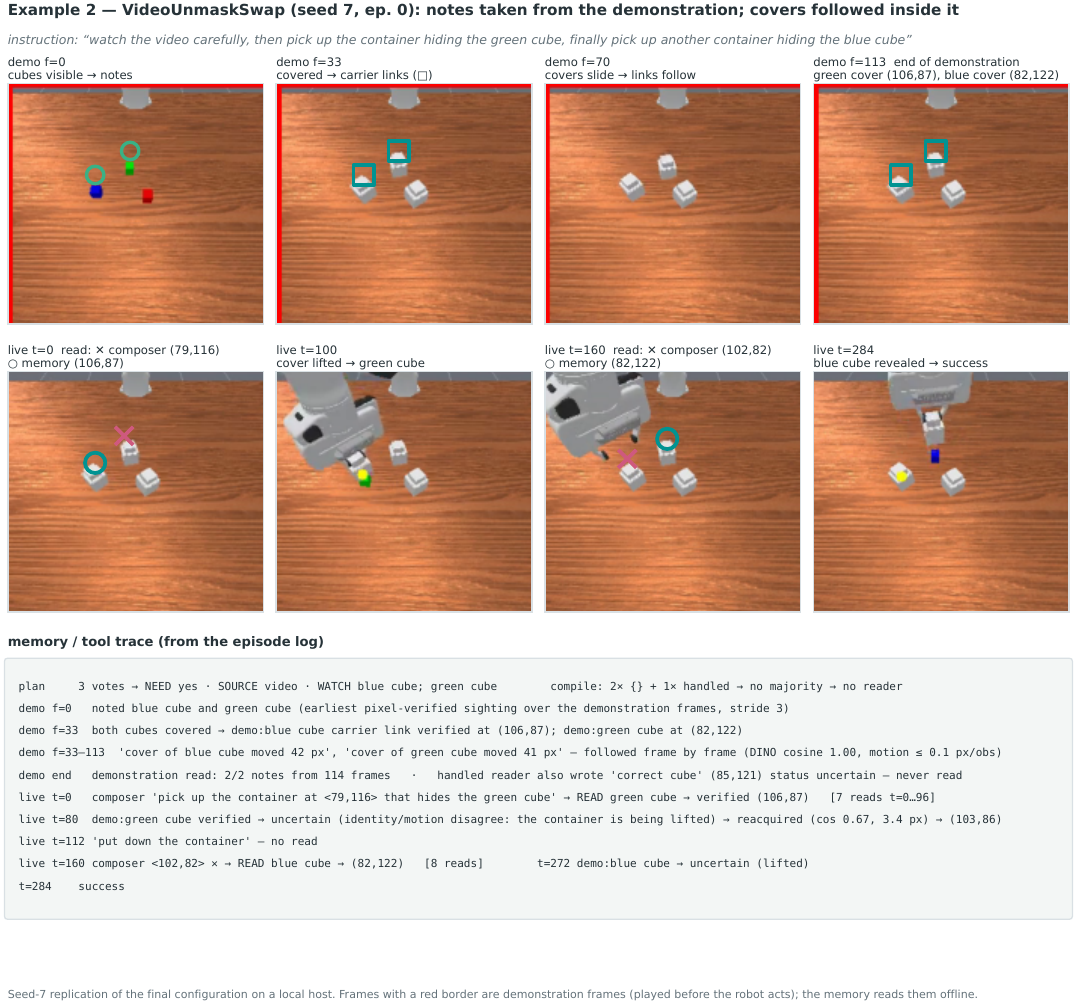}
\caption{Demonstration-conditioned notes and cover following on VideoUnmaskSwap.}
\label{fig:app_ex2}
\end{figure}

\begin{figure}[p]
\centering
\includegraphics[width=\textwidth]{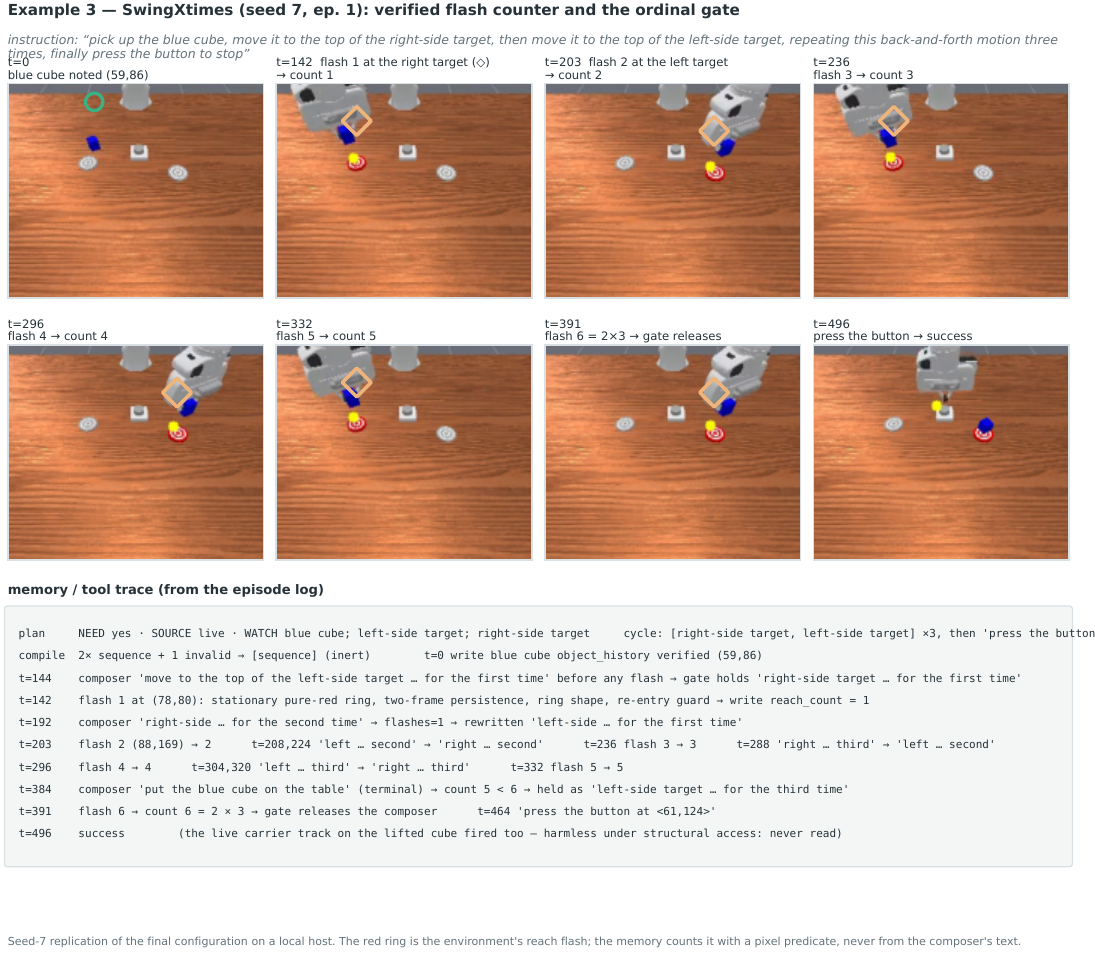}
\caption{Verified flash counting and ordinal access on SwingXtimes.}
\label{fig:app_ex3}
\end{figure}

\begin{figure}[p]
\centering
\includegraphics[width=\textwidth]{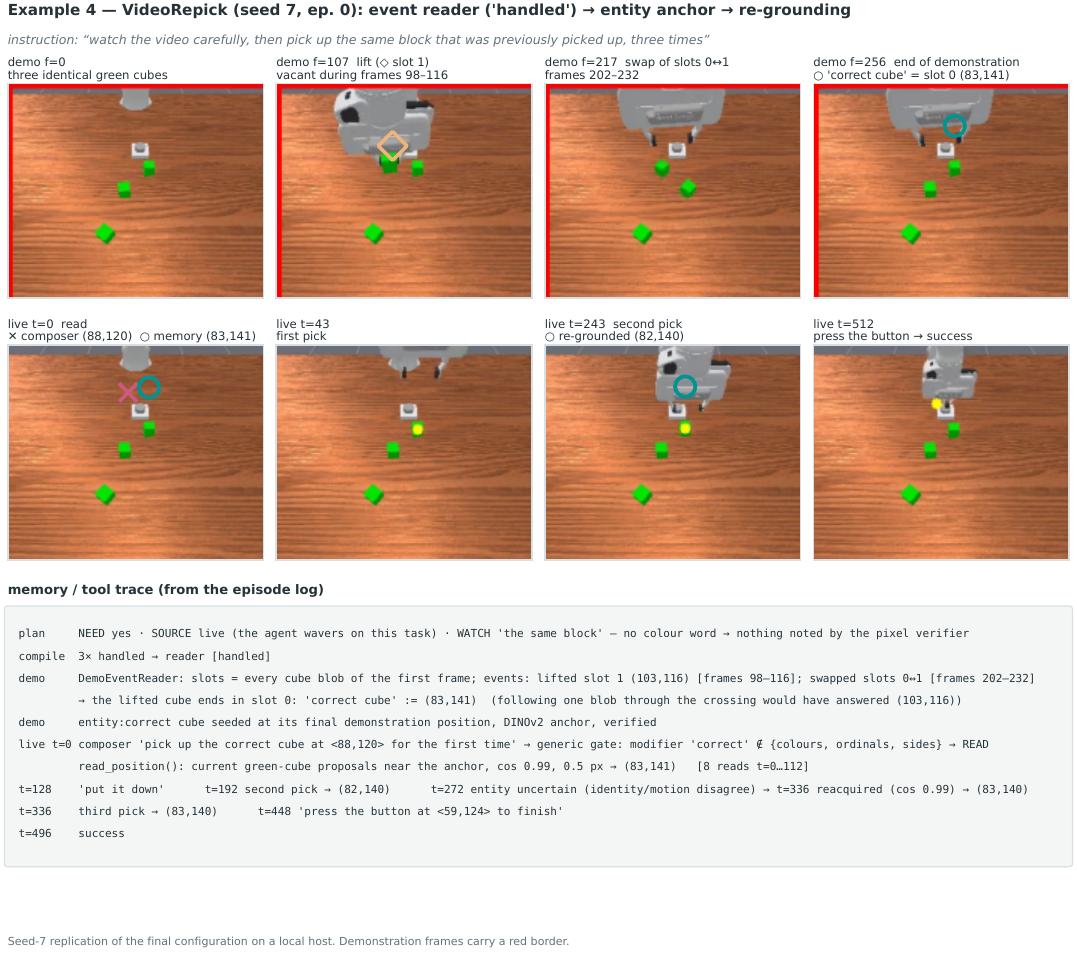}
\caption{Demonstration event reading, entity anchoring, and re-grounding on VideoRepick.}
\label{fig:app_ex4}
\end{figure}

\begin{figure}[p]
\centering
\includegraphics[width=\textwidth]{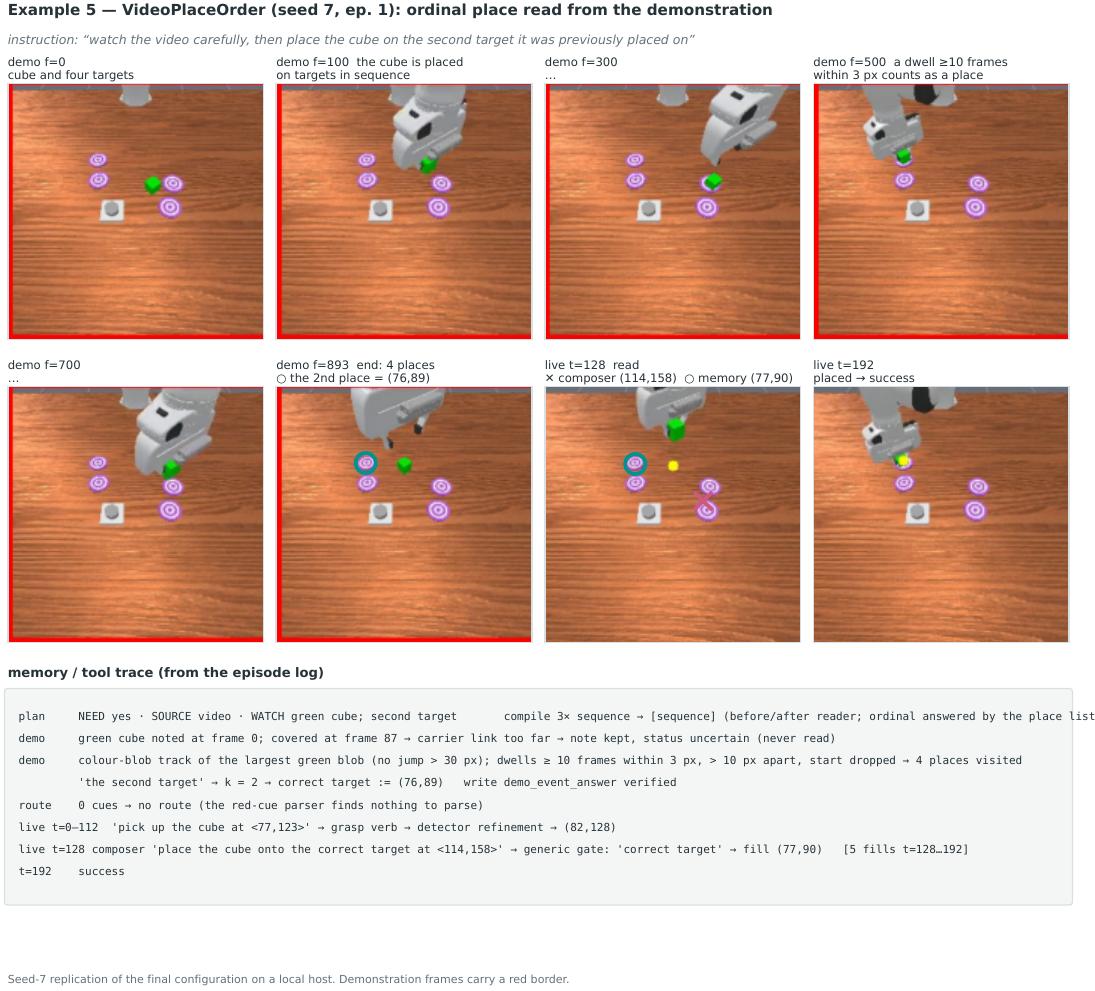}
\caption{Ordinal placement state extracted from the demonstration on VideoPlaceOrder.}
\label{fig:app_ex5}
\end{figure}

\begin{figure}[p]
\centering
\includegraphics[width=\textwidth]{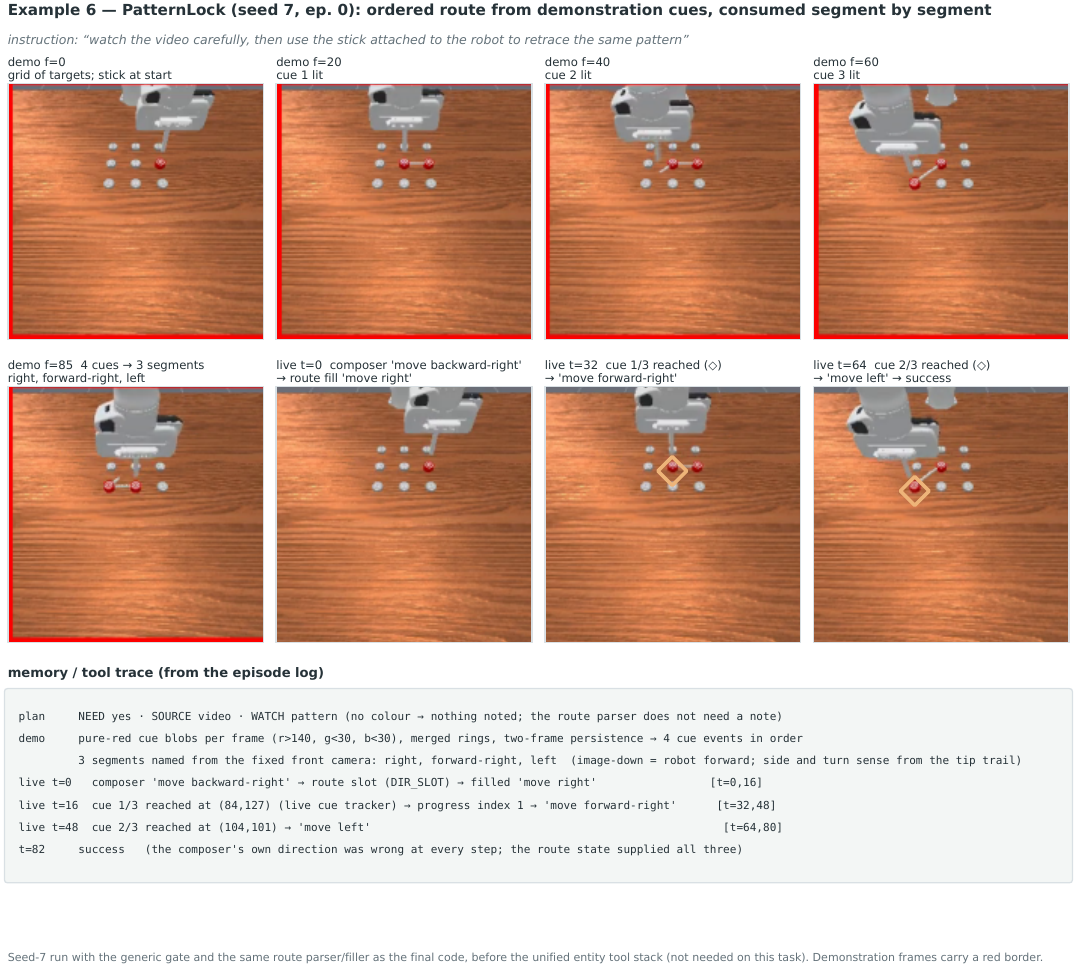}
\caption{An ordered route parsed from demonstration cues and retrieved segment by segment on
PatternLock.}
\label{fig:app_ex6}
\end{figure}

\begin{figure}[p]
\centering
\includegraphics[width=\textwidth]{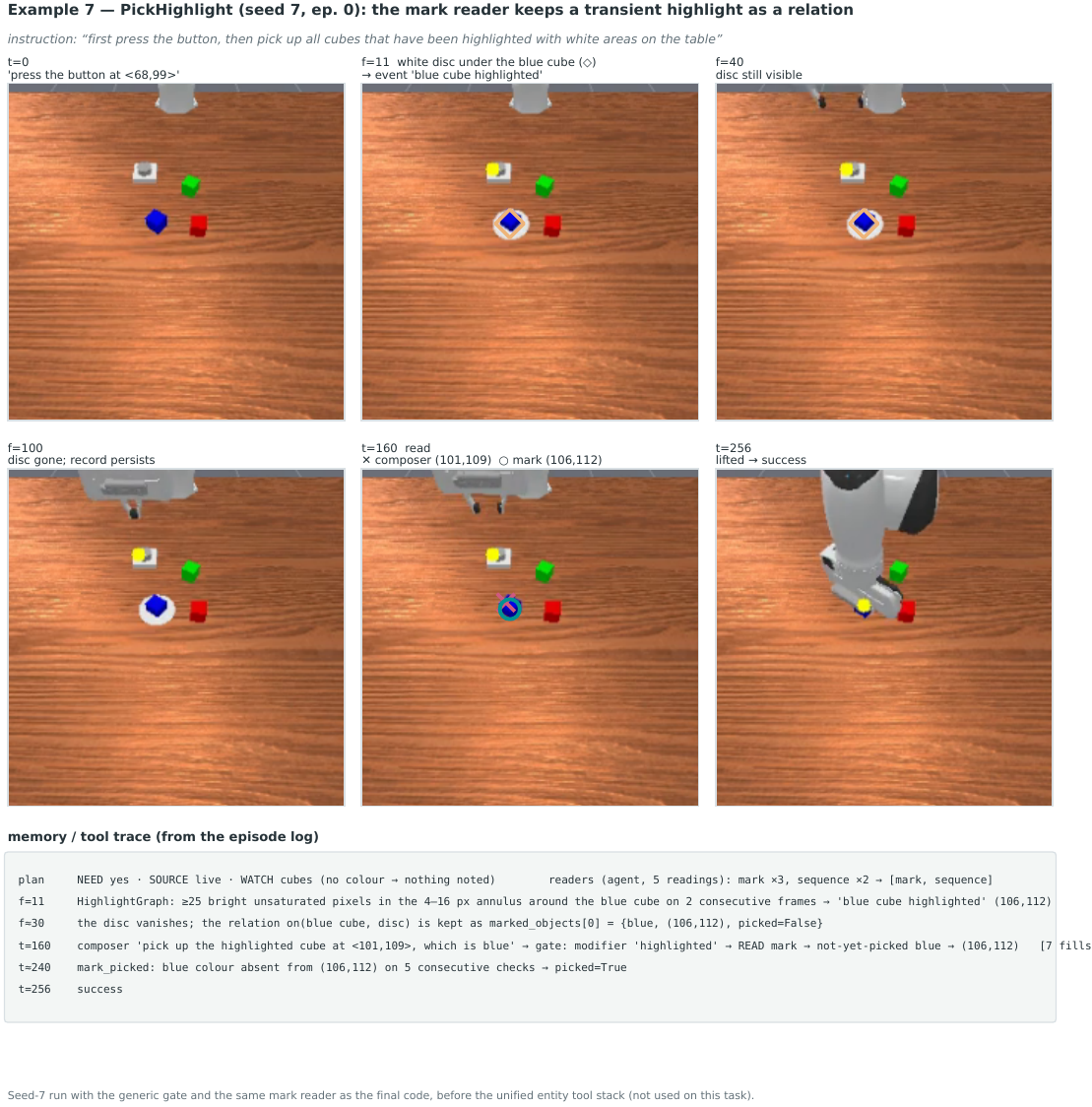}
\caption{A transient highlight maintained as a persistent entity--mark relation on PickHighlight.}
\label{fig:app_ex7}
\end{figure}

Figures~\ref{fig:app_ex1}--\ref{fig:app_ex7} provide complementary successful examples for the
binding, progress, reference, route, and mark mechanisms. Each combines rollout frames with a trace
transcribed from the episode log. Red borders denote demonstration frames, the yellow dot is the
coordinate sent to the VLA, and overlays mark memory coordinates ($\circ$), composer coordinates
($\times$), entity--cover bindings ($\square$), and event or cue locations ($\diamond$).
Figures~\ref{fig:app_ex1}--\ref{fig:app_ex5} are seed-7 replications of the final configuration on
a host where the mask tool was unavailable; the identity and geometric checks sufficed in the
illustrated reads. Figures~\ref{fig:app_ex6}--\ref{fig:app_ex7} use the same route and mark readers
as the final method but come from an earlier generic-gate run, before the unified entity tool stack,
which those tasks do not invoke.

\subsection{Remaining causal questions}

The current ablations establish the necessity and specificity of several typed variables and
the access rule. They do not isolate the PickHighlight mark reader, SAM, or current-view grounding,
nor do they uniquely decompose each failed episode between correct state, correct access, grounding,
and control. A mark-reader removal would test the one large unablated pathway, and a matched oracle
ladder---correct state, then oracle access, then oracle grounding---would provide finer attribution.
No missing cell is inferred from invocation frequency or internal verification.

\end{document}